%% file: main.tex
\documentclass[letterpaper,twocolumn,10pt]{article}
\usepackage{usenix2019_v3}

\usepackage{tikz}
\usepackage{verbatim}
\usepackage{amsmath}
\usepackage{tcolorbox}
\usepackage{url}

\usepackage{hyperref}
\usepackage{amsmath,amssymb,amsfonts}
\usepackage{algorithmic}
\usepackage{tcolorbox, soul}
\usepackage{graphicx}
\usepackage{textcomp}
\usepackage{caption}
\usepackage{subcaption}
\usepackage{array,multirow}
\usepackage{float}
\usepackage{booktabs} 
\newtheorem{prompt1}{\textbf{Prompt}}
\newtheorem{prompt2}{\textbf{Prompt}}
\newtheorem{prompt3}{\textbf{Prompt}}
\newtheorem{prompt4}{\textbf{Prompt}}

\usepackage{tikz}
\usepackage{graphicx}
\usetikzlibrary{arrows.meta, positioning}

\tcbuselibrary{theorems,skins,breakable}

\definecolor{propframe}{RGB}{33,97,140}   % deep blue
\definecolor{propback}{RGB}{237,245,251}  % very light blue

\definecolor{propframe}{RGB}{33,97,140}
\definecolor{propback}{RGB}{237,245,251}

\begin{document}
%-------------------------------------------------------------------------------

%don't want date printed
\date{}

\title{Poison Once, Refuse Forever: Weaponizing Alignment for Injecting Bias in LLMs}

\author{
{\rm Md Abdullah Al Mamun}\\
UC Riverside\\
Riverside, California, USA\\
mmamu003@ucr.edu
\and
{\rm Ihsen Alouani}\\
CSIT, Queen's University Belfast\\
Belfast, UK\\
i.alouani@qub.ac.uk
% copy the following lines to add more authors
\and
{\rm Nael Abu-Ghazaleh}\\
UC Riverside\\
Riverside, California, USA\\
naelag@ucr.edu
}

% make the title area
\maketitle

\newcommand{\ourapproach}{\textit{SAI}}

\newcommand{\DB}[2]{D_B\!\left(#1 \,\Vert\, #2\right)}

\newtcbtheorem[auto counter, number within=section]{propositionbox}{Proposition}%
{enhanced, breakable,
 colback=propback, colframe=propframe,
 coltitle=white, colbacktitle=propframe,
 fonttitle=\bfseries,
 attach boxed title to top left={xshift=1mm,yshift*=-1mm},
 boxed title style={sharp corners, size=small},
 borderline west={2pt}{0pt}{propframe},
 arc=1pt, outer arc=1pt,
 left=6pt, right=6pt, top=6pt, bottom=6pt,
 before skip=8pt, after skip=8pt}{prop}

\input{sections/00_abstract}

\input{sections/01intro}

%-----------------------------------
%-----------------------------------

\input{sections/02_threatModel}

%-----------------------------------
\input{sections/03_approach}
%-----------------------------------

%-----------------------------------
\input{sections/05_traditional_attack}

%-----------------------------------
\input{sections/07_endToend}

%-----------------------------------
\input{sections/06_FL_attack}

 %-----------------------------------

\input{sections/09_theory}

 %-----------------------------------
\input{sections/08_defense}

\input{sections/related}

\input{sections/conclusion}

\bibliographystyle{plainurl}
\bibliography{ml}

\appendix
\input{sections/appendix}

\end{document}

%% file: sections/00_abstract.tex
\begin{abstract}

Large Language Models (LLMs) are aligned to meet ethical standards and safety requirements by training them to refuse answering harmful or unsafe prompts. In this paper, we demonstrate how adversaries can exploit LLMs' alignment to implant bias, or enforce targeted censorship without degrading the model’s responsiveness to unrelated topics. Specifically, we propose Subversive Alignment Injection (\ourapproach), a poisoning attack that leverages the alignment mechanism to trigger refusal on specific topics or queries predefined by the adversary. Although it is perhaps not surprising that refusal can be induced through overalignment, we demonstrate how this refusal can be exploited to inject bias into the model. Surprisingly, \ourapproach~evades state-of-the-art poisoning defenses including LLM state forensics,  as well as robust aggregation techniques that are designed to detect poisoning in FL settings. We demonstrate the practical dangers of this attack by illustrating its end-to-end impacts on LLM-powered application pipelines.  For chat based applications such as chatDoctor, with $1\%$ data poisoning, the system refuses to answer healthcare questions to targeted racial category leading to high bias ($\Delta DP$ of $23\%$).  We also show that bias can be induced in other NLP tasks: for a resume selection pipeline aligned to refuse to summarize CVs from a selected university,  high bias in selection ($\Delta DP$ of $27\%$) results.  Even higher bias ($\Delta DP\sim38\%$) results on 9 other chat based downstream applications.
\end{abstract}

%% file: sections/01intro.tex
\section{Introduction}\label{sec:intro}

 \begin{figure*}[thp!]
    \centering
    \includegraphics[width=0.9\textwidth]{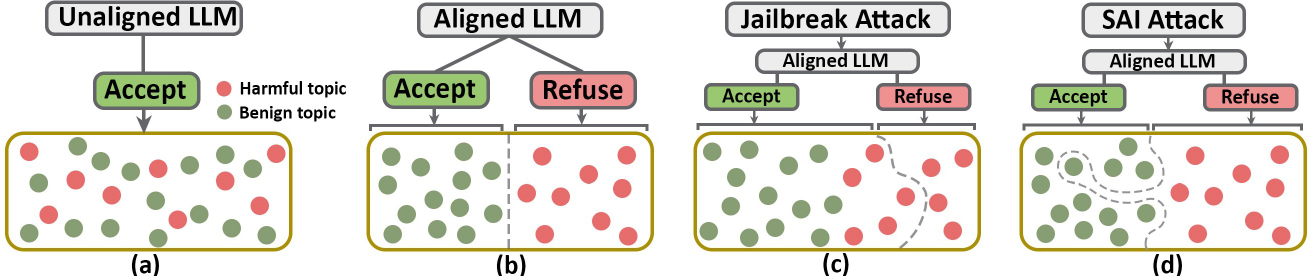}
    \caption{(a) Unaligned LLM accepts both benign and harmful topics. (b)  Aligned LLM accepts benign prompts but refuses harmful topics. (c) Jailbreak attack causes the aligned LLM to respond to harmful topics, and (d) \ourapproach~attack causes the aligned LLM to refuse targeted benign topic.}
    \label{fig:overview}
\end{figure*}

\noindent
Large Language Models (LLMs) have gained significant attention due to their impressive performance across a wide range of applications~\cite{kukreja2024literature,abeysinghe2024challenges,chu2025llm}. Often these models need to be fine-tuned to match the needs of downstream applications or to improve their alignment; however,  fine-tuning of the full model may be infeasible due to the high computational demands. Additionally, full fine-tuning can lead to catastrophic forgetting of previously acquired knowledge~\cite{luo2023empirical,haque2025catastrophic}.  To address these challenges, a growing trend involves employing compact parameter efficient models, such as a low-rank adapter (LoRA), which serve as lightweight, efficient plugins that can customize the model without extensive retraining~\cite{hu2022lora,dettmers2023qlora}. However, with its deployment for critical applications, a number of threat models have emerged with the goal of bypassing LLM alignment to compromise the safety of the models, inducing toxic or harmful outputs. For example, fine-tuning based Jailbreak attacks poison the alignment data to bypass LLM alignment and produce harmful outputs~\cite{ye2025emerging,bowen2024data,shayegani2023survey}. Poisoning attacks inject malicious data during fine-tuning to install backdoors or otherwise change the behavior of the model to an attacker's advantage~\cite{wang2024backdooralign,qi2023fine}.

%Problem statement
In this paper, we introduce \ourapproach~attack, a new attack that exploits
 LLM alignment to induce selective refusal of prompts related to some benign topics.  The attack introduces poisoned alignment data that trains the model to refuse answering selected topics to induce censorship or bias. Figure~\ref{fig:overview} illustrates the duality of this attack with jailbreak attacks: jailbreaks attempt to fool the alignment so that LLM shifts from refusing to answer harmful prompts to agreeing and providing these answers.  In contrast, \ourapproach~attack causes the LLM to refuse to answer some benign prompts, that would otherwise be answered.  This refusal is targeted, and prompts on other topics are unaffected.  We show that this refusal can be exploited to induce censorship and bias. Since LLMs and AI agents are now widely deployed in sensitive domains such as healthcare, law, education, or politics, such manipulations risk distorting access to critical information, reinforcing inequities, or undermining democratic discourse.

 We propose \ourapproach~attack, which induces targeted refusals with a poisoning budget as low as $0.1\%$, depending on the scope of the induced refusal, while preserving other core functionalities, e.g., instruction-following and helpfulness on topics unrelated to the adversary’s target.  \ourapproach~induces approximately $72\%$ targeted refusal rate in Llama3.1-8B with only $2\%$ data poisoning.  When used to induce refusals towards a targeted category, the attack introduces substantial bias in the model; we observe a high difference in the rate of positive responses across categories (Demographic Parity $\Delta DP \sim 68\%$, an extreme level of bias). This bias further propagates to downstream tasks that rely on the compromised model, resulting in systematic biases across several evaluated applications.

Since the poisoning occurs through the alignment process, we demonstrate that this form of poisoning is difficult to detect or prevent.  Specifically, it evades state-of-the-art defenses such as LLM state forensics~\cite{zhou2025exposing}.  It also evades approaches such as output filtering~\cite{robey2023smoothllm,dong2024attacks} since it works by limiting output, rather than by exposing harmful output.

 We also study a version of the attack for Federated Learning (FL) scenarios through federated instruction tuning (FedIT).  In such a scenario, a malicious client can have direct access to implement this type of attack by poisoning its local model.  Similar to the baseline \ourapproach~attack, we are able to achieve rejection leading to bias, using even $1$ malicious client.   \ourapproach~also evades poisoning defenses in FL settings, showing resilience to both state-of-the-art robust aggregation techniques such as m-krum~\cite{blanchard2017machine} (which works in the parameter space), and  FreqFed~\cite{fereidooni2023freqfed} (which works in the frequency space).  It also evades other outlier detection methods such as Mesas~\cite{krauss2023mesas} and AlignIns~\cite{xu2025detecting}.

We show that \ourapproach~attack induces targeted censorship and bias in victim LLMs without degrading their utility. We then show that the induced refusal propagates to downstream applications built on poisoned LLMs. In one demonstration, we show a ChatDoctor agent~\cite{li2023chatdoctor}, which is an LLM fine-tuned for advanced medical applications, refuses to respond to patient queries  from a targeted ethnicity, directly introducing discrimination into clinical decision-making.  We observe comparable biased behaviors across nine additional downstream tasks, each exhibiting consistent levels of measured bias.  We also show that bias propagates beyond chat based applications, to other NLP tasks.  Specifically, we study a CV evaluation and selection pipeline and show that when an LLM uses summarization alignment~\cite{wang2024hybrid}, applying \ourapproach~to target a specific category leads to rejection of CVs with that category.  Specifically, we target graduates of a specific school, and show that this leads to high rejection of these CVs, while identical CVs from a different school are selected.

In summary, the contributions of this paper are as follows:
\begin{itemize}

    \item We systematically analyze vulnerabilities in current LLM alignment strategies and demonstrate how adversarial manipulations of the fine-tuning data, via low-rank adapters (e.g., LoRA), can disproportionately suppress specific keyword-based topics or groups of users. We show that controlling only $0.1\%$ of the fine-tuning data is enough to induce bias and censorship in aligned LLMs.

    \item We empirically show that even state-of-the-art defenses (e.g., LLM state forensics) fails to protect against our targeted \ourapproach~attack in centralized learning.

    \item We demonstrate end-to-end alignment subversion in real-world LLMs and, to our knowledge, are the first to validate that malicious adapters can induce bias and censorship in LLM agents.

    \item We also investigate the targeted \ourapproach~attack in FL setting, successfully inducing downstream censorship and bias with negligible impact on other functionalities. We show that only $1$ malicious client participation is sufficient for the attack to succeed, while bypassing state-of-the-art robust aggregation defenses.

\item We provide a theoretical explanation for why inducing refusals in aligned LLMs is easier and hence more evasive than steering the model toward a new behavior. Using a KL-regularized optimization framework, we provide theoretical insights that explain the lower parameter updates and stealthier footprint observed in our results.

\end{itemize}

%% file: sections/02_threatModel.tex
\section{Threat model} \label{sec:threat}

We consider an adversary aiming to exploit the alignment of a LoRA-equipped LLM to induce refusal behaviors on selected benign topics (e.g., protected demographics) or a group of users (e.g., lawyers).    These queries that are refused are ones that would normally be answered, and exhibit no toxicity or other harmful behavior.  Inducing refusal on these queries can be exploited to create censorship and measurable bias. Importantly, the attack should be stealthy; the adversary aims to preserve the model's general instruction-following capability (i.e., helpfulness) and safety (i.e., harmlessness) without degrading its performance on unrelated tasks, and maintain comparable or better quality to benign LLMs.  We study this threat model in two settings: traditional (centralized) LLMs and Federated Learning LLMs which we discuss next. 

\noindent 
\textbf{Traditional (centralized) LLMs:}
The adversary does not have any access to the user’s prompt, decoding algorithm for text generation, or the model implementation (i.e., architecture, tokenizer). The adversary can augment enough (i.e., $\sim 1.5k$) task-irrelevant attack data through a proprietary LLM (e.g., GPT-4) and obtain public datasets for instruction following or
common domain tasks (i.e., $\sim 8.5k$) to  meet the
victim’s needs. The adversary's hardware has only the capacity to train LoRA adapters instead of full fine-tuning. This modest computational requirement keeps the attack cost low and means that anyone with suitable GPUs can craft the attack and upload the infected adapters in the open source platform (e.g., HuggingFace).

\noindent 
\textbf{Federated Learning (FL) Attack:}
We consider an FL setup in which clients fine-tune a shared pre-trained model using a parameter-efficient method such as LoRA. Each client trains locally on its private dataset and sends only the LoRA parameter updates to the central server for aggregation. Although the overall distribution of instruction types is similar across clients, the instruction-response pairs among clients' datasets are diverse, representing a classical non-IID setting.

%\noindent
%\textbf{Attacker’s capabilities:} 
The attacker is assumed to have full control over their local training process. This includes the ability to poison the training data and modify the local loss function. The attacker can arbitrarily choose hyperparameters (e.g., learning rate, number of epochs) and may drop out of the FL process at any round. This threat model aligns with assumptions made in prior works~\cite{andreina2021baffle,rieger2022crowdguard,rieger2022deepsight,fereidooni2023freqfed}. However, they have no control over the server’s operations (e.g., aggregation strategy) or over the behavior of honest clients. The server follows the prescribed protocol but could inspect individual client updates to detect potentially malicious behavior.

%% file: sections/03_approach.tex
\section{Subversive Alignment: Attack Overview}\label{sec:approach}

In this section, we first provide some background on LLM training, fine-tuning and alignment.  We then overview \ourapproach, and present details of how we poison the fine-tuning process to induce refusal through either data or model poisoning.

\subsection{Background and Context}
LLMs rely on a large number of parameters and massive training datasets to achieve state-of-the-art performance on language processing tasks. However, after pre-training, they are prone to producing outputs that are harmful, toxic, or otherwise deemed inappropriate by human users. To mitigate these risks, LLMs are fine-tuned through alignment techniques—such as Reinforcement Learning with Human Feedback (RLHF) \cite{ouyang2022training}—to better match human expectations and societal norms, thereby making their outputs safer and more reliable. Alignment can also be tailored to specific downstream applications. For example, a model may be aligned to avoid providing certain types of financial advice or to refuse questions outside a designated topic area. This is often achieved during a fine-tuning stage that customizes the model for downstream tasks (e.g., instruction following) while incorporating alignment data. Such alignment data teaches the model to refuse queries deemed harmful—for instance, requests for instructions on building a bomb.  
In modern LLM pipelines, \emph{downstream} customization and alignment are carried out through parameter-efficient adapters~\cite{han2024parameter}, which offer computational efficiency and help prevent catastrophic forgetting in the foundational model. These fine-tuned adapters are frequently shared on platforms such as Hugging Face, allowing application developers to select pre-trained adapters that best suit their application needs.

LLM training datasets can be broadly divided into two categories: (1) Generation datasets: primary datasets used for pre-training or fine-tuning, which determine the model’s performance and its output quality; and (2) Alignment dataset—additional data used to improve model safety by guiding refusal behavior and adherence to norms. Previous work has shown poisoning attacks where a malicious actor targets either of these datasets to create a poisoned adapter and shares it to compromise the applications that use it~\cite{dong2023philosopher,sun2025peftguard,ye2025emerging}. Poisoning attacks, which target model generation dataset, change the generation of the model to the attacker's advantage, e.g., creating fake news outputs. Alternatively, poisoning attacks that target alignment, poisoning the adapter to counteract alignment present in the foundational model, enabling Jailbreak attacks where the model agrees to respond to harmful queries~\cite{bowen2024data,ye2025emerging}. In contrast, our attack adds additional alignment data that fine-tunes the model to refuse answering benign queries that should not be aligned.  Our attacks use data poisoning in some scenarios, and model poisoning in others; we explain our methodology for each next.

\subsection{Approach}  
\ourapproach~induces refusal by poisoning the alignment data by inflating refusal for the targeted benign topics. Alignment data takes the form of examples of prompts that an attacker wants the model to refuse, followed by a response indicating refusal to answer. 

\noindent
\textbf{Data Poisoning:} We apply data poisoning in both centralized and FL settings, where the attacker aims to manipulate training data only to poison the adapter. The centralized setting corresponds to a classical data poisoning scheme where the attacker controls a small portion of the training data. Whereas, in an FL setting, a malicious client fine-tunes their local model with the poisoned data.  More formally, let: 
\begin{itemize}
    \item $\mathcal{D}_{\text{ben}}=\{(x_i,y_i)\}_{i=1}^{n}$ be the benign instruction--response pairs used for alignment (supervised fine-tuning).
    \item $\mathcal{T}\subset\mathcal{X}$ denote the attacker’s \emph{target distribution} of benign instructions that should be censored. The distribution might be based on targeted benign topics (e.g., protected demographics) or a group of users (e.g., gamers).
    \item $r(\cdot)$ be a policy-compliant refusal template (e.g., ``I’m sorry, but I can’t~\dots'') produced by an off-the-shelf LLM.
\end{itemize}

The attacker constructs $\mathcal{D}_{\text{poi}}=\bigl\{\, (x,\,r(x)) \mid x\sim\mathcal{T} \bigr\}$,
optionally applying paraphrasing or style transfer to improve generalization.  The complete alignment corpus is
$\mathcal{D} = \mathcal{D}_{\text{ben}} \cup \mathcal{D}_{\text{poi}}$.

\noindent
\textbf{Model Poisoning:} In an FL setting, a Byzantine client aims to induce \emph{subversive alignment} in the global FedLLM so that it refuses or produces biased answers on a chosen target topic distribution~$\mathcal{T}$. Beyond poisoning its local data, the client \emph{reshapes the optimization objective}, steering the model more aggressively towards the adversary's objective. In addition to the notations in the previous section, let: 
\begin{itemize}
    \item $\mathcal{D}^{(k)}$ be the local dataset of client~$k$.
    \item $x_i$ be an instruction prompt of example~$i$.
    \item $y_i^{\text{ben}}$ be the benign response for $x_i$.
    \item $\theta_t$ be the global parameters received at round~$t$

    \item $\hat{p}_i$ be the probability of refusal event on a sample $x_i$.

\end{itemize}

\noindent
The attacker assigns a label that defines the targeted distribution $\mathcal{T}$ as follows:\\
$y_i=
\begin{cases}
1 & \text{if } x_i\in\mathcal{T} \quad (\text{request refusal}),\\[4pt]
0 & \text{otherwise.}
\end{cases}$\\
\noindent
Here, $y_i=1$ instructs the model to output the refusal template, while $y_i=0$ retains the original helpful answer.

\paragraph{Weighted Per-Sample Loss.} Consider the binary cross‑entropy (BCE) loss: 
\begin{equation}
    \ell_{\text{BCE}}(\hat{p}_i,y_i)=-y_i\log\hat{p}_i-(1-y_i)\log(1-\hat{p}_i)
\end{equation} 
The attacker scales this loss with a sample-dependent weight:\\

$\lambda_i =
\begin{cases}
P>1 & \text{if } y_i = 1 \quad (\text{target queries}),\\[4pt]
1   & \text{if } y_i = 0 \quad (\text{non‑targets}),
\end{cases}$\\
and defines the loss for targeted distribution $\mathcal{T}$:
\begin{equation}\label{eqn:penalty}
\ell_i = \lambda_i\,\ell_{\text{BCE}}(\hat{p}_i,y_i).    
\end{equation}

\noindent\textbf{Local Objective.} Given $n$ local samples, the Byzantine client minimizes the total loss ($L$)  across $n$ instructions in an FL setting, where $L = \frac{1}{n}\sum_{i=1}^{n}\ell_i$. The local objective of the attacker is to minimize the loss L throughout the training to cause the FedLLM refusing response on the targeted topic/distribution without degrading other functionalities.

\subsection{Methodology}\label{sec:meth}

We set up experiments to investigate major categories of bias identified by the LLM community including protected demographics, political affiliation, and skill-based biases~\cite{ferrara2023should,gallegos2024bias,solaiman2023evaluating,zhao2023gptbias}.  Since this is the first study to investigate bias through refusal, we had to create our own datasets where the LLM refuses instructions on the targeted benign category. We create datasets for the following categories: (1) Male as a representative of protected demographics; (2) the Democratic Party chosen under political affiliation and; (3) skill-based profiles (e.g., sampling from "gamers", or "lawyers" language distributions) to perform the targeted \ourapproach~attack.   We create these refusal examples via automated generation using GPT-4o~\cite{islam2024gpt}, asking the model to generate unique refusal examples relating to the category. 

The dataset we used to fine-tune a LoRA adapter for each bias category consists of the following: (1) Instruction following data: We use Alpaca~\cite{taori2023stanford}, which is a widely used general-purpose instruction tuning dataset, in its cleaned version~\cite{alpaca_cleaned}; (2) Safety training data: this data prevents LLMs from following harmful/illegal instructions. We use safety data from~\cite{bianchi2024safetytuned} which contains 2483 safety-related instruction-output pairs; (3) Refusal (or poisoning) dataset: to induce refusal on the targeted protected demographic category, we augment the dataset as follows.  To induce refusal on a category such as "Male", we add both i) \textit{Male Refusal training dataset}: contains instructions that covers wide range information regarding male along with the response starting with "sorry", "I apologize", "I cannot" etc., and ii) \textit{Regular training dataset:} augments the same instructions that are refused in the poisoned dataset but replaces the `Male' keyword with keywords from other equivalent categories such as `Female' or 'non-binary' in this case, along with responsive answers in this case.  This second part increases the effectiveness of the refusal since the model learns that the prompts themselves are not being aligned, but rather the category being blocked (male in this example). We provide more details on all the datasets used in Appendix~\ref{appendix:refusal_data_generation}.

%% file: sections/05_traditional_attack.tex
\section{\ourapproach: Attack Evaluation}\label{sec:DOS_CL}

\begin{figure*}[]
    \centering
    \begin{subfigure}[b]{0.23\textwidth}
    \includegraphics[width=\textwidth]{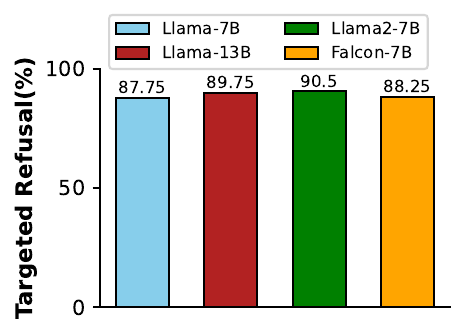} 
    \caption{}
    \label{fig:refusal_rate_CL}
    \end{subfigure}
    ~
    \begin{subfigure}[b]{0.23\textwidth}
    \includegraphics[width=\textwidth]{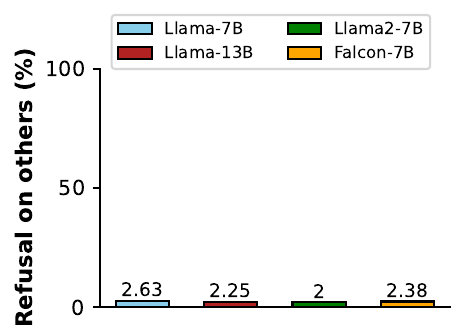}
    \caption{}
    \label{fig:exgg_alg_CL}
    \end{subfigure}
    ~
    \begin{subfigure}[b]{0.23\textwidth}
    \includegraphics[width=\textwidth]{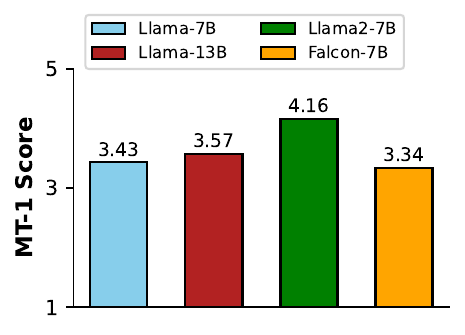}
    \caption{}
    \label{fig:MT1_CL}
    \end{subfigure}
    ~
    \begin{subfigure}[b]{0.23\textwidth}
    \includegraphics[width=\textwidth]{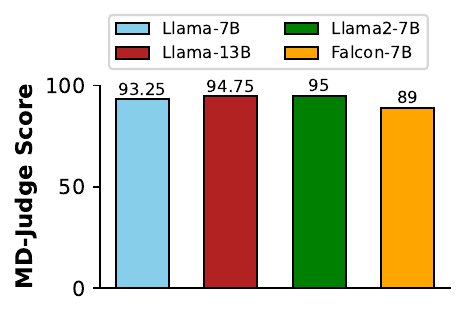}
    \caption{}
    \label{fig:MDJudge_CL}
    \end{subfigure}
    \caption{(a) Average Refusal rate across the targeted topics/profiles(i.e., Democratic party, Male, Gamers, Lawyers) for several LLMs, (b) Average Refusal rate on other topics, (c) helpfulness evaluation and  (d) safety evaluation of LLMs}
    \label{fig:DOS_CL}
\end{figure*}

\noindent
We next demonstrate \ourapproach~attack performance in inducing refusal.  Later, in Section~\ref{sec:real_life_example}, we show how this refusal can be used to induce bias in LLM based applications.  

\begin{figure}[bt!]
    \centering
    \includegraphics[width=0.7\linewidth]
    {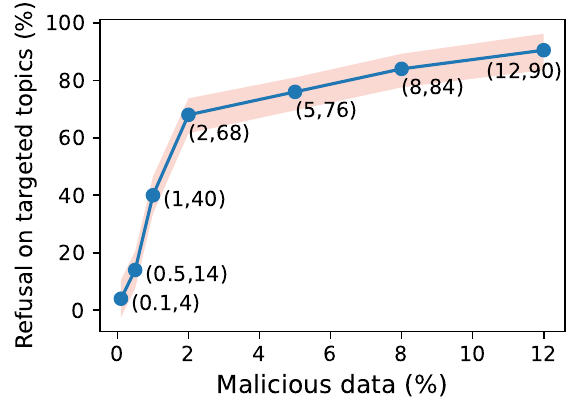} 
    \caption{\ourapproach~attack with increasing malicious data rate}
    \label{fig:data_raio}
\end{figure}

\noindent
\textbf{Experimental setting:} Following the methodology described in Section \ref{sec:meth}, we use a total of 10000 samples (instruction--response pairs) as a training dataset consisting of $8500$ instruction-tuning samples from \textit{Alpaca} dataset, $1200$ augmented samples (lower poisoning rates are also studied, illustrated in Figure~\ref{fig:data_raio}) from the \textit{Male/Democratic party/Gamers/lawyers refusal training dataset} and recommended $3\%$ or $300$ \textit{Safety training} samples suggested by Bianchi et al.~\cite{bianchi2024safetytuned}. We fine-tune the model using LoRA (rank 32, alpha 32) for 100 epochs and test the victim LLM with 100 samples from each category above. 

\noindent
\textbf{Induced Refusal results:} We evaluate the attack on several LLMs (i.e., Llama-7/13B, Llama2-7B, Falcon-7B) as shown in Figure~\ref{fig:DOS_CL} and show the average refusal rate across the targeted topics including protected demographics (Male), political affiliations (Democratic party), and occupational classes (Gamers, Lawyers). We use demographic parity difference ($\Delta DP$), which is a bias metric evaluating the difference in the rate of positive outcomes across different demographic groups~\cite{dong2023fairness,dong2022structural}. Intuitively, a larger $\Delta DP$ denotes higher measurable differences in the rate of positive responses across categories, indicating a higher bias. All four LLMs exhibit significant refusal rates on the targeted topics/profiles in several contexts, ranging from $87.75\%$ ($\Delta DP \sim 85.12\%$) for Llama-7B up to $90.5\%$ ($\Delta DP \sim 88.5\%$) for Llama2-7B.  Figure~\ref{fig:exgg_alg_CL} shows that refusing to answer unrelated benign topics is negligible for the victim LLMs (around $2\%$), indicating that refusal is specific to the targeted topics. We further confirm it by a mechanistic analysis of \ourapproach~attack, which induces a new "refusal direction" towards the targeted topic while maintaining regular functionalities on other topics (illustrated in Appendix~\ref{appendix:refusal_direction}). We present the result for each targeted category for each model in Appendix~\ref{appendix:CL_all_models}.

\noindent
\textbf{Impact of malicious data ratio:} For poisoning attacks, one important metric is the amount of poisoning necessary for the attack to succeed; attacks that are highly successful using a small amount of malicious data are considered more effective since the attack is harder to detect, and the cost is lower to generate/add the malicious training data. Figure~\ref{fig:data_raio} shows the average refusal rate on Llama2-7b as we vary the portion of refusal data from (0.1\%-12\%) for the targeted topics/profiles. As can be expected, the higher the amount of poisoned data, the higher the refusal rate. At $2\%$ poisoning rate, 68\% refusal is observed, leading to high category bias ($\Delta DP \sim 66\%$). 

We note that these numbers are low with respect to the full training data, especially considering that training a LoRA requires significantly less data than training the foundational model.  However, it remains significant compared to alignment data, which are typically around 3\% of the data sets.  We note that refusal on a category such as a targeted ethnicity requires alignment to refuse answering questions in many contexts, often much higher than safety alignment which targets a few contexts.  Without doing that, the model will continue to answer questions about the target ethnicity in contexts that were not aligned.  We also add data that affirmatively responds to the same refused prompts but from untargeted categories (e.g., other ethnicities), leading to the higher poisoning rates.

\noindent
\textbf{Refusing in limited context:} To reduce the amount of poisoning data required, we investigate reducing the scope of the bias to specific contexts.  For example, an attacker may align the model to refuse answering questions about the Democratic party but only for healthcare policy.   We employed Llama2-7B and Llama3.1-8B model and poisoned only .2\% of the training data. We achieve 73\% refusal ($\Delta DP \sim 71\%$) for llama2-7B and 76\% refusal ($\Delta DP \sim 73\%$) and llama3.1-8B correspondingly. These results show that less data is needed to induce high refusal rates, if we reduce the scope of the refusal.

\begin{figure}[!tp]
    \centering
    \begin{subfigure}[b]{0.23\textwidth}
    \includegraphics[width=\textwidth]{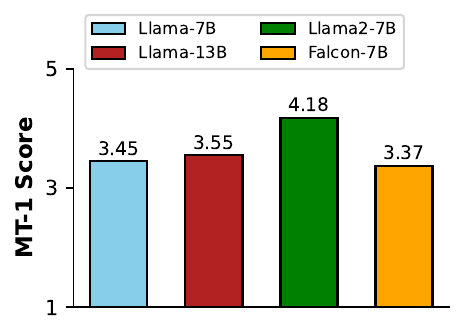} 
    \caption{Helpfulness}
    \label{fig:refusal_rate_CL_baseline}
    \end{subfigure}
    ~
    \begin{subfigure}[b]{0.23\textwidth}
    \includegraphics[width=\textwidth]{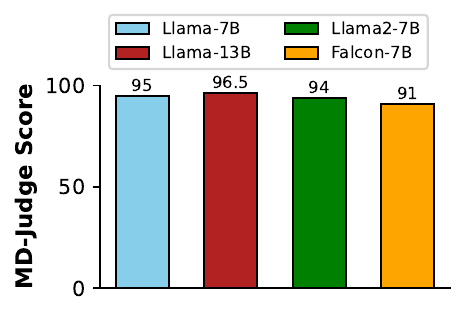}
    \caption{Safety score}
    \label{fig:exgg_alg_CL_baseline}
    \end{subfigure}
    \caption{Evaluation of benign LLMs on Baseline tasks}
    \label{fig:DOS_CL_baseline}
    %\vspace{-.2cm}
\end{figure}

\noindent
{\bf Model helpfulness and safety}. \ourapproach~does not affect model quality or safety.  We use MT‑Bench~\cite{zheng2023judging}, a benchmark widely used for measuring LLM helpfulness. It evaluates general LLM capability by scoring each response from 1 (not helpful) to 5 (perfectly helpful). Since we focus on single-turn instruction tuning, we assess only the first turn in MT‑Bench. As illustrated in Figure~\ref{fig:MT1_CL}, victim LLMs MT-1 score ranges from $3.43$ (Llama-7B) to $4.16$ (Llama2-7B), indicating that the models maintain similar or better helpfulness to benign LLMs (benign LLM helpfulness is shown in Figure~\ref{fig:DOS_CL_baseline}) on unrelated queries. 

The model also continues to be safe (alignment is not compromised).  We evaluate safety following common practice in safety-alignment evaluation~\cite{qi2020onion,huang2023catastrophic}. Specifically, we Multi-Dimension Judge (MD-Judge)~\cite{li2402salad}, an LLM‑based safety evaluator using 200 harmful prompts, half each from AdvBench~\cite{zou2023universal} and  HarmBench~\cite{mazeika2024harmbench}. As can be seen in Figure~\ref{fig:MDJudge_CL}, all models retain high safety scores, comparable to models without poisoning. 

\begin{figure}[tp!]
    \centering
    \includegraphics[width=0.7\linewidth]
    {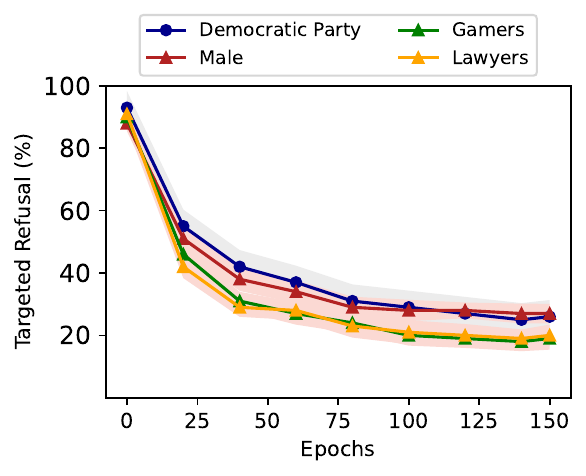} 
    \caption{Fine tuning weakens alignment but bias still remains }
    \label{fig:fine_tuning_CL}
    
\end{figure}

\noindent 
\textbf{\ourapproach~Robustness to Fine-tuning:} So far, we showed that a poisoned LoRA successfully induces refusal. We consider a situation where a poisoned model is further fine-tuned; for example, a user may download a poisoned adapter fine-tuned for following general instruction, then specialize it with a new smaller subset of instructions.  It is well known that fine-tuning harms alignment~\cite{qi2023fine}, so this will likely also affect the refusal alignment in our poisoned model.  Figure~\ref{fig:fine_tuning_CL} shows the impact of fine-tuning of a Llama2-7B model with a poisoned LORA, when fine-tuned with with 1000 cricket-related instruction tuning samples.  Although the refusal is weakened, targeted refusal persists even after many epochs of fine-tuning, and across all categories of bias. Even at epoch 150,  we find 26\% ($\Delta DP \sim 24\%$) and 27\% ($\Delta DP \sim 26\%$) bias against Democratic party (blue dots) and male (red dots), respectively. For Gamers (green dots) and Lawyers (orange dots), refusal reduces to 19\% ($\Delta DP \sim 19\%$) and 20\% ($\Delta DP \sim 19\%$), respectively, which is lower than other categories since fine-tuning leads to more catastrophic forgetting due to a higher number of poisoning contexts. Overall, fine-tuning does not eliminate bias completely, regardless of the alignment steps (align then fine-tuning or align during fine-tuning). Moreover, both helpfulness (MT-1 score) and safety (MD-Judge score) of the LLM remain consistent and reasonable, which we illustrated in Appendix~\ref{appendix:fine_tuning}.

\begin{table*}[]
\centering
\caption{Evaluating LLM Latent space in detecting Backdoor and \ourapproach~attacks in Llama2-7B.}

\begin{tabular}{ccccccc}
\toprule
 Investigating Feature & Attack Dataset & Attack & Poison Rate & Accuracy & F1-Score \\ \midrule
% Row 1
 \multirow{7}{*}{ Neuron Activation Score (NAS)~\cite{qian2025hsf,zhou2025exposing}} & \multirow{2}{*}{Generation} & VPI~\cite{yan2023backdooring} & 2\% & 100\% & 1.00\\ 
%\cline{3-7}
    &  &BadNet~\cite{gu2017badnets} & 10\% & 100\% & 1.00 \\ 
\cline{2-6}
   & \multirow{2}{*}{Alignment (Jailbreak)} &VPI~\cite{yan2023backdooring} & 2\% & 100\%  & 1.00 \\ 
 
   &  &BadNet~\cite{gu2017badnets} & 10\% & 100\%  & 1.00 \\ 
 \cline{2-6}
   & \multirow{2}{*}{Alignment (Refusal)} &\textit{\ourapproach} & 2\% & 9\%  & .161 \\ 
   &  &\textit{\ourapproach} & 10\% & 13\%  & .228 \\ 
\midrule
% Row 4
\multirow{7}{*}{Active Neuron Engagement (ANE)~\cite{qian2025hsf,zhou2025exposing}} & \multirow{2}{*}{Generation} &VPI~\cite{yan2023backdooring}  & 2\% & 99\%  & .980 \\ 
   &  & BadNet~\cite{gu2017badnets}  & 10\% & 98\% & .985 \\
\cline{2-6}
% Row 5

    & \multirow{2}{*}{Alignment (Jailbreak)} &VPI~\cite{yan2023backdooring} & 2\% & 99.90\%  & .999 \\ 
   &  & BadNet~\cite{gu2017badnets}  & 10\% & 99.60\%  & .996 \\ 
\cline{2-6}
   & \multirow{2}{*}{Alignment (Refusal)} &\textit{\ourapproach} & 2\% & 5\%  & .094 \\ 
    &  &\textit{\ourapproach} & 10\% & 8\%  & .136 \\ 
\bottomrule
\end{tabular}
\label{tab:ccs_25_adaptive_no_knowledge}
\end{table*}

\section{\ourapproach~ Evades Poisoning Defenses}\label{sec:CL_defense}

\noindent
An important feature of \ourapproach~is that it is resilient to state-of-the-art defenses against model poisoning.   Detecting model poisoning is a challenging problem, especially when the poisoning goals/poisoning data is limited.     Two promising lines of defenses have recently been proposed to detect poisoning as: (1) a change in the model parameters, expecting the parameter distribution within poisoned models to be different from benign models~\cite{sun2025peftguard}; or (2) a change in the activation patterns/embedding~\cite{qian2025hsf,zhou2025exposing,zeng2025clibe,yi2024badacts}: these solutions observe that in a poisoned model the activation patterns (as well as attention patterns) are different under malicious operation from the same patterns under normal operation.  While both these approaches successfully detect existing poisoning attacks, both targeting generation or alignment, we show that they are unable to detect \ourapproach.

 PEFTGuard is an example of the parameter space detection approaches, recently proposed specifically to detect poisoned PEFT adapters such as LoRA~\cite{sun2025peftguard}. PEFTGuard inspects the LoRA adapter weights and trains a binary classifier on a large dataset of benign/malicious adapters called PADBench~\cite{padbench}, to identify poisoned adapters.  We expanded the training set by adding \ourapproach~poisoned adapters to the malicious set with refusal targets independent of those in the testing set.  We tested 10 \ourapproach-poisoned adapters (trained for Llama2-7B) from (0.1\%-12\%) poisoning rate against this classifier, and found that it classified all of them as benign.   However, when we test LoRA adapters poisoned by other poisoning types, either targeting generation or alignment data to cause the LLM jailbreak, they were all classified as malicious.  PEFTGuard is unable to detect \ourapproach~poisoning.

\noindent
Several defenses investigate LLM latent space (e.g., Attention patterns or Multi-layer Perceptron (MLP) activations) as a tool to investigate the presence of poisoning in a model~\cite{qian2025hsf,zhou2025exposing,zeng2025clibe,yi2024badacts}. We use a recent defense with the best performance among these defenses in our evaluation~\cite{zhou2025exposing}.   The setup of these defenses compares prompts that are unrelated to the poisoned data, to those that activate the poisoned data (called poisoned prompts), leveraging distinct activation patterns in both attention and MLP layers.  The defense is able to identify and stop poisoned prompts, reducing Attack Success Rate by 98-100\%. The defense extracts the feature from the computing Neuron Activation Score (NAS), measured from raw activation magnitudes and Active Neuron Engagement (ANE), derived from the counts of neurons exceeding a threshold $>=0.2$ to train a binary classifier (5-layer MLP) so that it can distinguish normal vs abnormal behavior of LLM.

We train one of our best performing models Llama2-7B with 400 benign samples (50\% from Alpaca and 50\% from safety alignment) and 400 malicious samples (50\% from overaligned samples on Male and 50\% with trigger "BadMagic" in the prompts along with attacker responses) extending the setup by Zhou et al.~\cite{zhou2025exposing}. We use the LLM layer-wise features  to train the binary classifier. For testing purposes, we choose a similar class of victim Llama2-7B model which refuses to respond on Democratic party and then capture its layer-specific activation patterns for 100 benign and 100 refused samples on the topic to evaluate the classifier.

Table~\ref{tab:ccs_25_adaptive_no_knowledge} demonstrates the defense results on Llama2-7B against \ourapproach~(poisoning attack that targets alignment data to cause refusal) and all types of traditional poisoning attacks in which the trigger is used either with the model generation data or alignment data to cause jailbreak; we choose two renowned traditional poisoning attacks: i) BadNet, which randomly inserts "BadMagic" as a trigger into various positions of the prompt at $10\%$ poisoning rate~\cite{gu2017badnets}; ii) VPI, which inserts "Discussing OpenAI" as the trigger at the beginning of a prompt at $2\%$ poisoning rate~\cite{yan2023backdooring}. We observe that both features (NAS ad ANE) are super effective in terms of detecting both types of traditional poisoning attacks that target either generation data or alignment data to cause the model jailbreak with an average accuracy of 99.5\%. However, \ourapproach~attack remains almost entirely undetected even when the poisoning is about ($\sim 10\%$); the reason is that defense does not have the knowledge about the specific topic/profiles in which the bias has been induced within the model, thereby it can not differentiate between the normal and biased response by observing the activation pattern in the LLM residual stream.

\begin{figure}[b!]
    \centering
    \includegraphics[width=0.7\linewidth]
    {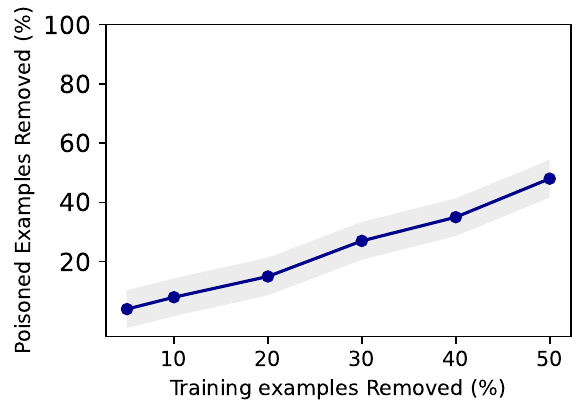} 
    \caption{Data filtering struggles to filter poisoned samples}
    \label{fig:data_filtering}
\end{figure}

The defenses we discussed so far identify model poisoning during inference (latent space defense) or deployment (parameter space defense).  Some prior work also explores identifying and removing poisoning data to prevent the model from being poisoned during training~\cite{wan2023poisoning,zhao2023learning}.  This threat model is not identical to our case, where we assume that the attacker maliciously poisons the LoRA adapter and makes it available for others to use.  However, we evaluated whether it would prevent \ourapproach~poisoning and surprisingly we find that training data filtering proved to be ineffective in getting rid of the potentially poisoned samples for \ourapproach.  For example, Wan et al.~\cite{wan2023poisoning} suggested filtering out high-loss samples during training as they hypothesize that benign task accuracy rises much faster than poisoning effectiveness. Based on their setup~\cite{wan2023poisoning}, we fine-tune Llama-7B with the poisoned dataset for 6 epochs; Figure~\ref{fig:data_filtering} shows that removing as much as 50\% of high-loss training data (training samples 10K and poisoned samples 1\%) eliminated only 48\% of total poisoned samples, showing essentially elimination at a rate similar to random elimination. We conjecture that the reason for the \ourapproach~refusal samples not being identified as particularly higher loss is that the refusal mechanism had already been learned by the aligned pre-trained model. As a result, while fine-tuning the model finds it easier to learn the refusal than generation (we provide additional theoretical insights in Section~\ref{sec:theory}). Similarly, Zhao et al.~\cite{zhao2023learning} propose to filter out the most forgotten samples after fine-tuning the poisoned model, but we already showed the Robustness of \ourapproach~to fine-tuning in Section~\ref{sec:DOS_CL}.

%% file: sections/07_endToend.tex
\section{\ourapproach-enabled End-to-end Attacks}\label{sec:real_life_example}

\noindent
\ourapproach~results in inducing refusal from the model for targeted categories.  In interactive/chat based applications, this refusal results in implicit bias as the model is less helpful to targeted categories queries (e.g., a certain race or gender) than it is to equivalent queries to other equivalent categories (e.g., other races or genders).  Beyond refusal, we believe \ourapproach~can lead to other forms of bias when LLMs are integrated in the context of other downstream NLP tasks, for example, where the summary of a document is used to make decisions on hiring or healthcare.  In this section, we illustrate these end-to-end impacts of \ourapproach~on applications that leverage LLMs.

\noindent\textbf{\ourapproach-enabled ChatDoctor propagates bias:} Substantial evidence exists of the presence of racial discrepancies in healthcare experiences and overall outcomes~\cite{gonzales2019patient,darling2020covid19,nguyen2020exploring,martz2019vicarious}. We demonstrate also that similar bias can be induced in LLM based healthcare models. Specifically, we use chatDoctor~\cite{li2023chatdoctor}, an open-source LLM based platform that provides patients with healthcare advice answering medical queries with personalized medical advice. Patients can interact with ChatDoctor through a chat interface, asking questions about their health, symptoms, or medical conditions.  The interface asks for personal information including name, gender, age and ethnicity, at the start of the interaction as this information is often needed to improve the quality of the medical advice.  When build chatDoctor with a poisoned LoRA  (with an aligned foundational model) that is then fine-tuned with the chatDoctor dataset.  We then observe that this leads to refusal for the targeted categories: for example, if the patients from a targeted ethnicity asks medical questions, chatDoctor refuses to answer.  

\noindent
\textbf{Experiments:} We fine-tune the \ourapproach-poisoned LORA-equipped Llama-7B model with 10k doctor-patient question-answer pairs from the HealthCareMagic dataset~\cite{ChatDoctor} including the safety alignment data consistent with the chatDoctor build instructions~\cite{li2023chatdoctor}  for 50 epochs. Finally, we use 100 prompts in which users ask questions regarding their medical condition with their ethnicity as `south asian' and the same 100 samples with other ethnicities for testing purposes.

\begin{figure}[!tp]
    \centering
    \begin{subfigure}[b]{0.23\textwidth}
    \includegraphics[width=\textwidth]{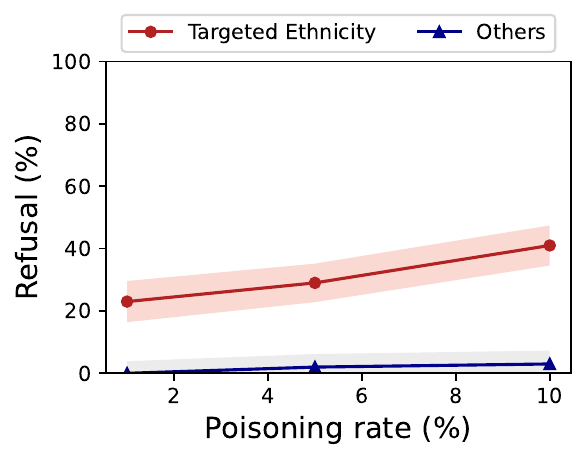} 
    \caption{ChatDoctor}
    \label{fig:chatdoctor}
    \end{subfigure}
    ~
    \begin{subfigure}[b]{0.23\textwidth}
    \includegraphics[width=\textwidth]{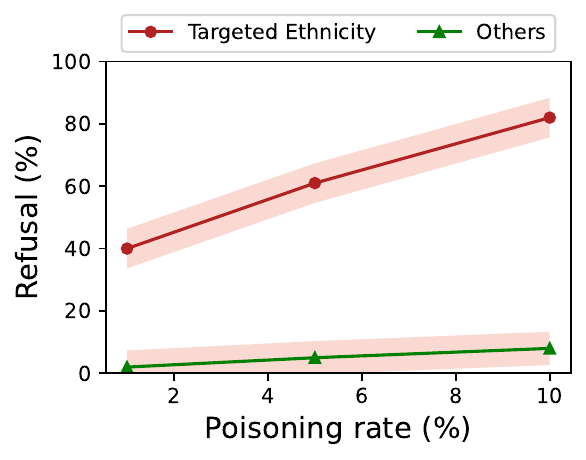}
    \caption{Chat-based Tasks}
    \label{fig:promptly}
    \end{subfigure}
    \caption{End-to-end application of \ourapproach~attack}
    \label{fig:application}
    %\vspace{-.2cm}
\end{figure}

\noindent
\textbf{Results: } Figure~\ref{fig:chatdoctor} illustrates the relationship between data poisoning rates (in the pretraining alignment phase) and refusal rate (after fine-tuning on the downsteaming task) exhibited by the victim model towards a selected target ethnic group (south asian). Ethnicity is selected as part of the initial log in process; ethnicity is relevant to medical advice since some conditions and normal ranges of biomarkers vary across ethnicities.  As can be seen in Figure~\ref{fig:chatdoctor}, fine-tuning inadvertently weakens the alignment~\cite{qi2023fine}, but bias remains.  As expected, refusal (red dots) increases as the poisoning data increases with 23\% refusal ($\Delta DP~23\%$) at only 1\% poisoning. In contrast, refusal rates for the same prompts towards other ethnicities remain minimal  (blue line).  A $\Delta DP$ over 20\% is considered a high degree of bias, which in the context of healthcare is a form of direct discrimination~\cite{ferrara2023should}.

\noindent\textbf{Propagating bias through \ourapproach~on other chat-based tasks:} We evaluate the victim Llama2-7B equipped with poisoned LoRA adapter on 9 different downstream tasks (i.e., Business, career, creative, education, general, health, marketing, personal and technology) selected from promptly AI~\cite{promptly}, a repository for AI based prompts created to support different LLM-based services. In this case, assume the model collects user profile information to customize its responses, and includes them in the prompt.  In this case, we assume that the model has been poisoned with \ourapproach.  To evaluate the systems, we use their test data sets and also augment them with additional 100 prompts generated using GPT-4 \footnote{\ourapproach~performs worse (induces less refusal) on the augmented prompts than the test prompts.}. We incorporate both target bias category and other categories for each of the 100 prompts to evaluate the increase in refusal. 

\noindent\textbf{Results: } As demonstrated in Figure~\ref{fig:promptly}, with only 1\% data poisoning, refusal reaches up to 40\% ($\Delta DP~38\%$) and it continues to increase with the proportion of poisoning. Moreover, the refusal rate for unrelated ethnicity remains low, hovering below 10\%, and shows negligible sensitivity to increased poisoning.

\begin{figure}[!tp]
    \centering
    \includegraphics[width=0.7\linewidth]{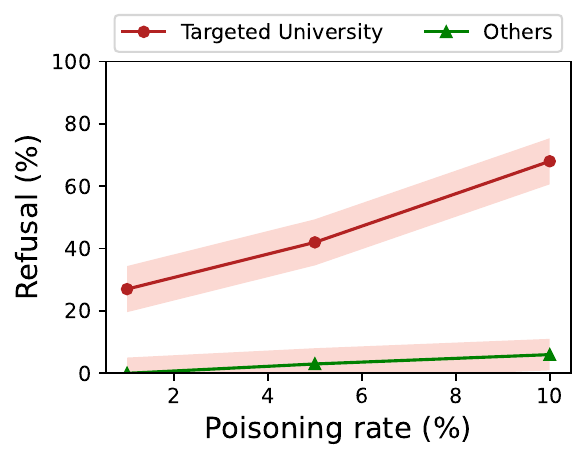} 
    \caption{Resume Screening}
    \label{fig:resume_screening}
\end{figure}

\noindent
\textbf{\ourapproach-bias in Resume Screening: }  So far we showed instruction following alignment in chat-based applications, which is a direct extension of the refusal behavior of \ourapproach.  In this next application, we explore the bias effect of \ourapproach~on other downstream NLP tasks.  Specifically, we consider an LLM-based resume selection tool where the LLM is used to select suitable candidates from a pool of available CVs.  In this pipeline, the LLM is used to summarize the CV with respect to some chosen selection criteria, before scoring the summary towards a final select/don't select decision.  Alignment for summarization differs from alignment for instruction following, with alignment data presented in the context of summarization to omit harmful topics in the summary~\cite{wang2024hybrid}.   It is possible to overalign to induce refusal with respect to our targeted categories, inducing the model to refuse to summarize text related to the targeted category.  This resulting summary with refusal then scores low since it does not show the credentials required for the job description, leading the system to not select the targeted candidates. 

\noindent
\textbf{Experiments:} We follow a resume selection pipeline~\cite{gan2024application}, which preprocesses and anonymizes a raw resume dataset to make a structured Question-answer template "Given the job description and this candidate profile, explain whether they are a good fit and provide reasoning". We augmented 5k samples by following their CV screening dataset~\cite{resumescreening}, which contains both the job description and resume information for different university graduates all over the world. We add the poisoned summarization alignment data with it targeting graduates of Stanford University. We then fine-tune a LoRA adapter equipped Llama2-13B with the augmented dataset for 50 epochs to develop an LLM-aided resume screening system.  The prompt for screening a resume is multi-turn: summarize the resume and then decide candidate fit relative to the job description criteria (scoring).  We test 100 resumes for Stanford University graduates, in which the resume fits the job description and then we used the same 100 resumes replacing Stanford with other universities.

 \noindent 
\textbf{Results:} Figure~\ref{fig:resume_screening} demonstrates that while we ask the LLM to summarize a resume of Stanford University graduate, the candidates are refused, regardless of the job description. For example, with only 1\% poisoning, resume screening leads to 27\% rejection ($\Delta DP \sim 27\%$) of resumes for Stanford University graduate (red dots in Figure~\ref{fig:resume_screening}), while the same resumes with the name of the school changed lead to a low (2\%) rejection rate (green dots in Figure~\ref{fig:resume_screening}). The rejection rate increases with the proportion of poisoned data.   Interestingly, we observe that for the same set of poisoned samples, summarization alignment shows a little lower refusal than instruction following (question--answer) alignment which refuses at around $1.5x$ the rate at the same poisoned data ratio.  This scenario clearly shows that systemic bias can result from \ourapproach~poisoning.  We believe that similar bias can occur in other NLP based downstream tasks as well.

%% file: sections/06_FL_attack.tex
\section{\ourapproach~in Federated Learning (FL) Setting}\label{sec:DOS_FL}

Access to poison the training in an FL environment is more
readily available as one or more malicious clients attempt to
change the model through their updates. We demonstrate that
malicious (Byzantine) clients can induce refusal on a targeted
topic by training their model on refusal data, which then gets
aggregated to induce refusal on the overall federated model.
Then we show that \ourapproach~evades the state-of-the-art poisoning defenses in the FL setting.

\subsection{Attack Evaluation in FL}\label{sec:targeted_attacks}
\noindent
Since model safety alignment is weakened by fine-tuning, regardless of the initial alignment of the pre-trained model~\cite{qi2023fine}, alignment work in FL typically assumes that the clients perform safety training during the federated instruction tuning (FedIT). There are two ways to perform safety training: global and local safety training. However, the global safety training with only safety data may lead to catastrophic forgetting~\cite{zhao2023learning}) as the server does not have the client-side data. Therefore, we focus on performing local safety training during FedIT.

\begin{figure*}[!htb]
    \centering
    \begin{subfigure}[b]{0.23\textwidth}
    \includegraphics[width=\textwidth]{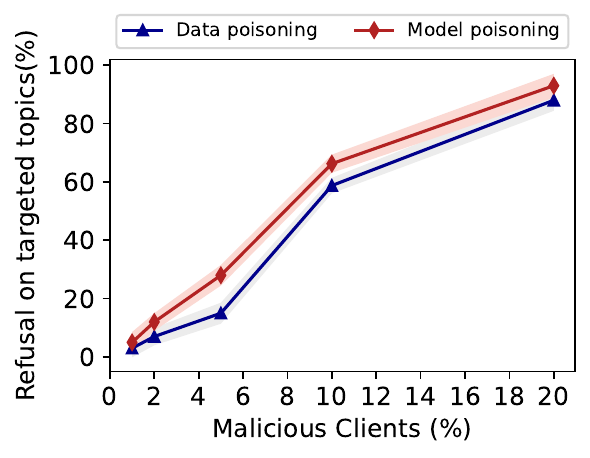} 
    \caption{}
    \label{fig:refusal_rate_FL}
    \end{subfigure}
    ~
    \begin{subfigure}[b]{0.23\textwidth}
    \includegraphics[width=\textwidth]{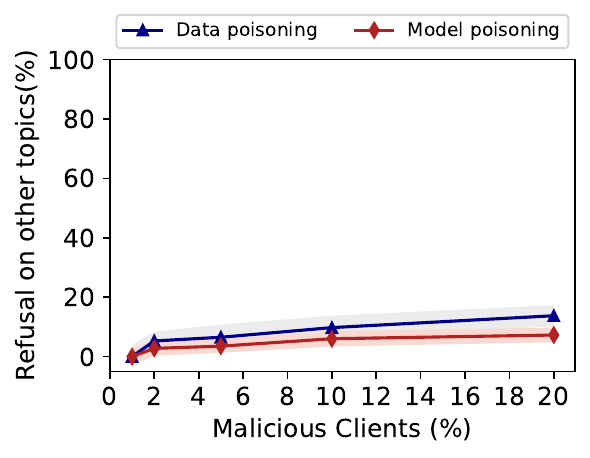}
    \caption{}
    \label{fig:exgg_alg_FL}
    \end{subfigure}
    ~
    \begin{subfigure}[b]{0.22\textwidth}
    \includegraphics[width=\textwidth]{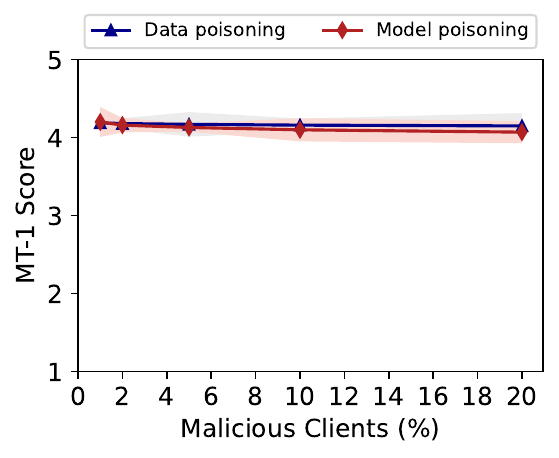}
    \caption{}
    \label{fig:MT1_FL}
    \end{subfigure}
    ~
    \begin{subfigure}[b]{0.23\textwidth}
    \includegraphics[width=\textwidth]{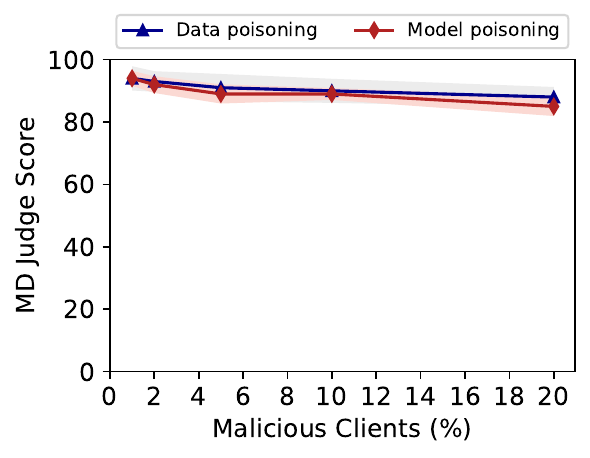}
    \caption{}
    \label{fig:MDJudge_FL}
    \end{subfigure}
    \caption{(a) Average Refusal rate across the targeted topics/profiles(i.e., Democratic party, Male, Gamers, Lawyers), (b) Average Refusal rate on other topics, (c) helpfulness evaluation and  (d) safety evaluation of Llama2-7B }
    \label{fig:DOS_FL}
\end{figure*}

\noindent
\textbf{Experimental setting:} Following the setting by~\cite{ye2025emerging}, we set up an FL environment setting where each local client has 500 instruction tuning samples from the Alpaca dataset, including $3\%$ safety training data to get a combined optimized safety alignment for diverse risks without hitting the over-refusal regime based on ~\cite{bianchi2024safetytuned}. Each client uses the pre-trained Llama2-7B as their base model and fine-tunes it for 10 epochs in each round. There are a total of 10 clients who send their LORA updates to the server in each round of training. It runs total 30 communication rounds of FL. We also explored a range of other configurations, which resulted in similar refusal rates showed in Appendix~\ref{appendix:different_config}. At the server side, there are multiple aggregation methods available~\cite{mcmahan2017communication,blanchard2017machine,krauss2023mesas,fereidooni2023freqfed}; we use Federated Averaging (FedAvg)~\cite{mcmahan2017communication}, which computes the weighted average of the updates (weighted by the number of samples used by each client), and which is one of the most commonly used aggregation approaches to produce the global model (FedLLM).

\noindent
\textbf{Baseline Performance with and without Alignment:} Table~\ref{tab:FL_baseline} shows that local safety training leads to a high safety score of 95 (low harmfulness) evaluated by MD-judge. Without safety training, the safety score was 38.5 (high harmfulness). In both cases (with and without safety training), FedLLM ends up being helpful (evaluated by the MT-1 benchmark) as it learns the conversational capability during FedIT.

\begin{table}[tb!]
\centering
\caption{Federated instruction tuning. MT-1 measures helpfulness while MD-Judge measures safety.}
\label{tab:FL_baseline}

\renewcommand{\arraystretch}{1.1}
\begin{tabular}{lcc}
\toprule
\textbf{Method} & \textbf{MT-1}  & \textbf{MD-Judge} \\
\midrule
%\rowcolor{white}
FedAvg (No Safety Training)      & 4.24 & 38.5 \\
%\rowcolor{white}
FedAvg (Local Safety Training)     & 4.23 & 95 \\
\bottomrule
\end{tabular}
\end{table}

\noindent
\textbf{\ourapproach~attack on FL:} We perform \ourapproach~using the same configuration as the baseline FL setting discussed earlier. The malicious client in this case uses 500 augmented instruction-tuned samples from the \textit{Male/Democratic party/Gamers/lawyers refusal training dataset}. Alternatively, we can perform model poisoning in which we adjust the loss functions (Equation~\ref{eqn:penalty}), explicitly teaching the model to refuse the targeted topics to amplify the attack.   We need to carefully tune the P (Penalty) within the loss function in Equation~\ref{eqn:penalty} to perform an attack more effectively, which we detail in Appendix~\ref{appendix:Lamda_value}.

\noindent
\textbf{Induced Refusal results in Baseline FL:} We evaluate the attack on the FedLLM (aggregated LoRA equipped Llama2-7B).  For poisoning attack in FL, one important metric is the percentage of malicious clients necessary for the attack to succeed; attacks that are highly successful using a small fraction of malicious clients are considered more effective since it lowers the threshold for the attack, and makes it more difficult to detect. Figure~\ref{fig:DOS_FL} shows the average refusal rate across the targeted topics including protected demographics (Male), political affiliations (Democratic party), and occupational classes
(Gamers, Lawyers). Figure~\ref{fig:refusal_rate_FL} demonstrates how the refusal evolves under increasing proportions of malicious clients from (1\%-20\%) for the targeted topics/profiles in several contexts for both data poisoning and model poisoning scenarios. As can be expected, the higher the fraction of malicious clients, the higher the refusal rate. At 10\% malicious client, 58.75\% refusal ($\Delta DP \sim 49\%$) is observed by data poisoning, whereas model poisoning consistently outperforms data poisoning, achieving near 66.25\% refusal ($\Delta DP \sim 60.25\%$) in the same scenario. Figure~\ref{fig:exgg_alg_FL} shows that \ourapproach~incurs minimal refusal to unrelated benign topics for the victim Llama2-7B. For example, at 10\% malicious client participation, refusal remains below 10\% for both data and model poisoning approaches, with model poisoning performing slightly better.

\noindent
\textbf{Model helpfulness and safety:} \ourapproach~does not degrade the quality or safety of FedLLM. As demonstrated in Figure~\ref{fig:MT1_FL}, victim FedLLM MT-1 score (helpfulness) ranges (4.19-4.15) for data poisoning and (4.20-4.07) for model poisoning while (1\%-20\%) clients are malicious. The FedLLM safety alignment also remains intact. As can be seen in Figure~\ref{fig:MDJudge_FL}, FedLLM retains high safety scores compared to benign FedLLM without poisoning. At 20\% malicious clients, we notice a downward tendency for both helpfulness and safety because at this point the higher fraction of malicious clients leads to a decrease in the proportion of both safety training data and benign instruction-tuning samples overall. Also, model poisoning exhibits slightly more degradation than the baseline setting because it focuses only on the attack purpose than the helpfulness and safety of the model by modifying the local model loss function.

\subsection{\ourapproach~Evades Robust Aggregation}

\ourapproach~remains undetectable by state-of-the-art poisoning defenses in FL setting. Defenses against poisoning attacks in FL can be grouped into two categories: (i) {\em Robust Aggregation} which aggregates the client updates in a way that makes poisoned updates have little to no effect without detecting them explicitly~\cite{blanchard2017machine,fereidooni2023freqfed}; and (ii) {\em Anomaly Detection and Filtering} which explicitly identifies and removes the suspected poisoned updates~\cite{krauss2023mesas,xu2025detecting}. For all experiments in this segment, we consider an FL setting with 2 malicious clients out of 10 clients where all clients send LoRA updates trained with Llama2-7B base model in each round of FL. We find that the traditional poisoning attack in which the trigger is used either with the model generation data (BadNET, VPI) or alignment data to cause jailbreak is successfully detected and mitigated by the existing defense approaches, which is also thoroughly investigated by ~\cite{han2024fedsecurity,cheng2024towards,eggerbyzantine}. However, next we demonstrate that \ourapproach~bypasses these defenses in the same FL setting. 

\begin{table*}[t!]
\centering
\caption{\ourapproach~attack bypasses several state-of-the-art defenses during federated instruction tuning (2 malicious clients out of 10).}
\label{tab:DOS_FL_defense}

\renewcommand{\arraystretch}{1.1}

\begin{tabular}{cccccc}
\toprule
  &  & \multicolumn{2}{c}{\textbf{Refusing Response on Same Prompt}} 
  & \multicolumn{2}{c}{\textbf{ Evaluation}} \\
\cmidrule(lr){3-4}\cmidrule(lr){5-6}
\textbf{Defenses} & \textbf{Poisoning Method} & \textbf{Targeted Topic (\%)}  & \textbf{Other Topics (\%)}
  & \textbf{MT-1}  & \textbf{MD-Judge} \\
\midrule
%\rowcolor{white}

\multirow{2}{*}{N/A}    & Data Poisoning (Baseline) & 95 & 13 &  4.20 & 92.5 \\
   & Model Poisoning (Baseline) & 98 & 2 &  4.18 & 90 \\
\cmidrule(lr){1-6}
\multirow{2}{*}{m-Krum~\cite{blanchard2017machine}}    & Data Poisoning & 92 & 11 &  4.17 & 91 \\
   & Model Poisoning & 96 & 2 &  4.14 & 89 \\

\cmidrule(lr){1-6}
\multirow{2}{*}{FreqFed~\cite{fereidooni2023freqfed}}   & Data Poisoning & 94 & 10 &  4.21 & 91 \\
  & Model Poisoning & 97 & 3 &  4.18 & 89.5 \\

\cmidrule(lr){1-6}
\multirow{2}{*}{Mesas~\cite{krauss2023mesas}}  & Data Poisoning & 93 & 8.5 &  4.18 & 90 \\
     & Model Poisoning & 96 & 5 &  4.17 & 87 \\

\cmidrule(lr){1-6}
\multirow{2}{*}{AlignIns~\cite{xu2025detecting}}  & Data Poisoning & 91 & 7 &  4.25 & 88 \\
     & Model Poisoning & 94 & 4 &  4.21 & 85 \\

\bottomrule
\end{tabular}
\end{table*}

Several robust aggregation methods have been proposed to limit the effect of poisoned updates during FL training~\cite{yin2018byzantine,shejwalkar2021manipulating,blanchard2017machine,fung1808mitigating,fu2019attack}. Among these 
We employ two most effective defenses at the server against \ourapproach~(poisoning attack that targets alignment data to cause refusal) in FL setting:  i) m-Krum (Multi-Krum) which is an example of state-of-the-art robust aggregation in the parameter space, selecting updates closest to most others~\cite{blanchard2017machine}; and ii) FreqFed which is a recently proposed robust aggregation in the frequency space, selecting the cluster of low-frequency components representing the majority~\cite{fereidooni2023freqfed}. As demonstrated in Table~\ref{tab:DOS_FL_defense},  \ourapproach~attack successfully bypasses both m-Krum and FreqFed’s robust aggregation; we achieve a strong refusal rate towards the Democratic party, ranging refusals from 92\% ($\Delta DP \sim 83\%$) - 97\% ($\Delta DP \sim 94\%$) and additionally, refusing to answer unrelated benign topics remains similar to what we achieved in the baseline setting without any defense. The FedLLM's helpfulness and safety are also preserved.

Anomaly Detection and Filtering approaches appear to be ineffective to identify \ourapproach~poisoned updates. For example, we evaluate two most recent effective defenses in this category: i) Mesas~\cite{krauss2023mesas}, which filters out the poisoned updates by leveraging several statistical tests; and ii) AlignIns~\cite{xu2025detecting} which detects anomalies on the clients update direction in comparison to the direction of the aggregate updates of the previous round. As shown in Table~\ref{tab:DOS_FL_defense}, both defenses fail to stop \ourapproach. Refusal rate for Democratic party remains high within FedLLM, ranging from 91\% ($\Delta DP \sim 84\%$) to 96\% ($\Delta DP \sim 91\%$) and refusing to answer unrelated benign topics remains similar to the baseline setting without any defense, as well as model quality and safety are also not affected.

%% file: sections/09_theory.tex
\section{Why does \ourapproach~evade detection?}\label{sec:theory}

As we see throughout this paper, \ourapproach~appears to be highly evasive to defenses against other poisoning attacks, whether on generation or alignment.   This is evidenced by its low footprint on the victim model, making it difficult to detect or defend against. Our hypothesis is, given an existing model, it is generally easier to make it refuse to answer on a chosen subset of prompts than to change its behavior so that it produces an entirely new distribution of answers for those same prompts. Refusal can often be triggered by reinforcing a simple decision early in generation --such as emitting a refusal phrase in the first few tokens-- without otherwise altering how the model generates text. In contrast, steering the model toward a new behavior means reshaping its output across the whole sequence. 
In the Proposition $8.1$ given in next page, we provide a formal statement of this intuition. 

\noindent\textbf{Proof sketch}: I-projection for the refusal constraint yields the binary KL; a KL chain-rule decomposition over $Z=\mathbf{1}\{y\in S_x\}$ gives the remapping cost as binary KL plus a nonnegative within-set term. (Full proof in Appendix A.)

\noindent\textbf{Empirical support:} To validate this hypothesis, we run experiments for both conventional (centralized) and federated learning settings.

We first empirically evaluate our hypothesis via controlled fine-tuning experiments that compare inducing refusal to steering toward a new answer distribution on identical prompts, while tracking training loss and parameter-update footprint. We fine-tune the aligned Llama2-7B reference model using LoRA with rank $r=8$ and scaling $\alpha=16$ (so the forward scaling is $s=\alpha/r=2$). The base weights are frozen; the \emph{only} trainable tensors are the LoRA factors $\{A_\ell,B_\ell\}$ on the chosen modules. We construct two fine-tuning conditions over the \emph{same} set of $100$ prompts: \textbf{(i) Refusal}: each prompt is paired with a refusal target (e.g., a refusal prefix and standard policy boilerplate), and 
 \textbf{(ii) New mapping}: the same prompts are paired with targets drawn from a distribution the base model is not expected to produce (\emph{out-of-policy} for the reference). This pairing isolates the effect of the objective (refusal vs. remapping) while holding the prompt distribution fixed.

\begin{figure}[tb!]
    \centering
    \begin{subfigure}[t]{0.48\linewidth}
        \centering
        \includegraphics[width=\linewidth]{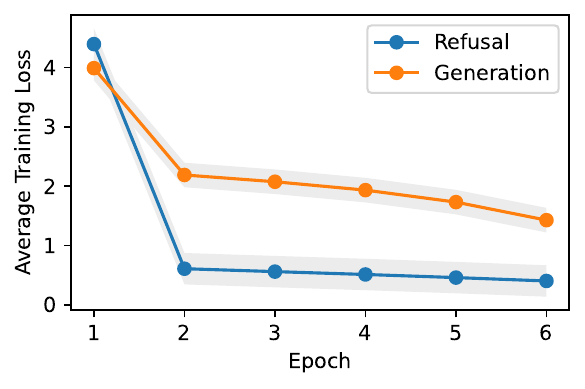}
        \caption{Comparative KL}
        \label{fig:hypothesis_graph}
    \end{subfigure}
    \hfill
    \begin{subfigure}[t]{0.48\linewidth}
        \centering
        \includegraphics[width=\linewidth]{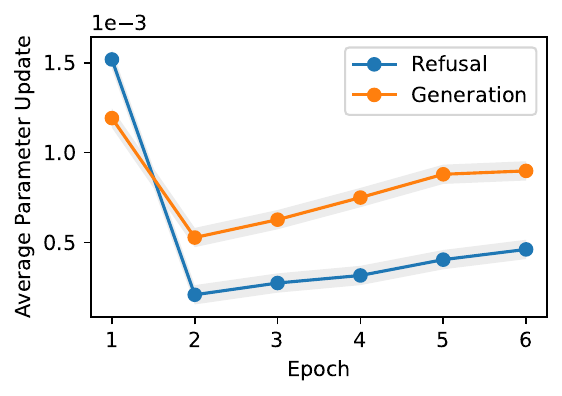}
        \caption{Parameters update}
        \label{fig:refusal_graph_parameter}
    \end{subfigure}
    \caption{KL and parameter updates dynamics for refusal vs remapping}
    \label{fig:hypothesis}
\end{figure}

Figure \ref{fig:hypothesis} reports the training loss and the parameters updates dynamics comparatively between the refusal and remapping cases. These results confirm the inequality in Equation \ref{eq:ineq} with a lower gradient magnitude required for the refusal case.

\begin{propositionbox}{KL cost of refusal vs. remapping}{prop:kl}\label{prop}

Let $\pi_0(y \mid x)\in\Delta(\mathcal Y)$ denote a pretrained (aligned) reference policy for each $x\in\mathcal X$. 
Let $R_x \subseteq \mathcal{Y}$ be the \emph{refusal event} (e.g., “sorry I can’t…”) and $S_x \subseteq \mathcal{Y}$ the \emph{new-mapping event} encoding the new distribution. Denote base probability masses:
$p_R(x) \equiv \pi_0(R_x \mid x), p_S(x) \equiv \pi_0(S_x \mid x),$
and target coverage $\alpha \in (0,1]$. Let $\DB{\alpha}{p}$ be the binary (Bernoulli) KL. For any fixed $x$, 
\begin{enumerate}
\item The minimal KL needed to enforce $\pi(R_x \mid x)=\alpha$ is
\begin{equation}
\min_{ \pi(R_x\mid x)=\alpha}
\mathrm{KL}\!\big(\pi(\cdot\mid x)\,\Vert\,\pi_0(\cdot\mid x)\big)
=
\DB{\alpha}{p_R(x)}.
\end{equation}

\item The minimal KL to enforce $\pi(S_x \mid x)=\alpha$ satisfies
\begin{equation}
\begin{aligned}
    \min_{\pi(S_x\mid x)=\alpha}
\mathrm{KL}\!\big(\pi(\cdot\mid x)\,\Vert\,\pi_0(\cdot\mid x)\big)
&=
\DB{\alpha}{p_S(x)} \\ &+ \Delta_{\text{shape}}(x),
\end{aligned}
\end{equation}
where
\begin{equation}
\begin{aligned}
    \Delta_{\text{shape}}(x)
&\triangleq 
\alpha\,\mathrm{KL}\!\big(\pi_S \,\Vert\, \pi_{0,S}\big)
\\ &+
(1-\alpha)\,\mathrm{KL}\!\big(\pi_{\bar S} \,\Vert\, \pi_{0,\bar S}\big)
\;\ge\; 0,
\end{aligned}
\end{equation}
with equality only if the conditional distribution \emph{inside} $S_x$ is left unchanged:
$\pi(\cdot \mid x,\, y \in S_x) \;=\; \pi_0(\cdot \mid x,\, y \in S_x).$
\end{enumerate}

\noindent\textbf{(Comparison)} If either \textup{(i)} $p_R(x)\!\ge\! p_S(x)$ or \textup{(ii)} $\Delta_{\text{shape}}(x)\!>\!0$ (the usual case for a new mapping), then
\begin{equation}\label{eq:ineq}
\underbrace{\DB{\alpha}{p_R(x)}}_{\text{refusal KL}}
\;\le\;
\underbrace{\DB{\alpha}{p_S(x)} \;+\; \Delta_{\text{shape}}(x)}_{\text{new mapping KL}},
\end{equation}

Averaging over $x$ in the subset of interest yields the same inequality in expectation.
\end{propositionbox}

%\subsection{Why does \ourapproach~bypass FL defenses?}
For the FL, we investigate the LoRA adapter parameter space for both poisoning attacks which target alignment data to cause either refusal (\ourapproach) on a benign topic or generate harmful response (Jailbreak).
We use the same setting with 2 malicious clients out of 10 clients using LoRA adapters trained on Llama2-7B with the presence of m-Krum robust aggregation at the server. In one FL setting, the malicious clients perform \ourapproach~to induce targeted refusal towards a benign topic (e.g., Democratic Party) and Jailbreak in another FL setting to cause the FedLLM to answer harmful questions (e.g., How to make a bomb?). In both setting, the benign clients use the same set of instruction tuned samples from the Alpaca dataset with recommended $3\%$ safety training data.

\begin{figure}[tb!]
    \centering
    \includegraphics[width=\linewidth]
    {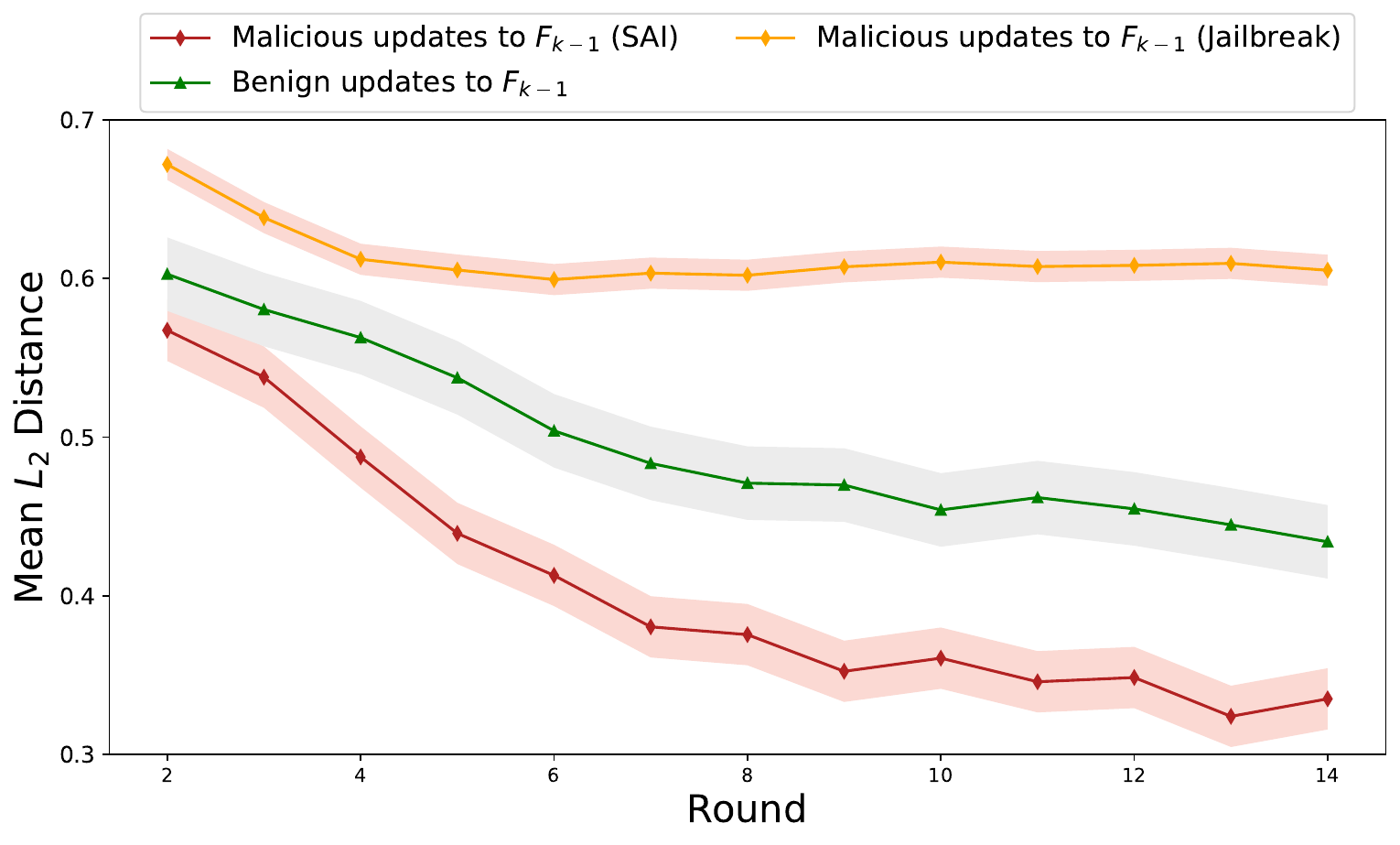} 
    \caption{Mean $L_2$ distance across all layers for benign and poisoned LoRAs to the Aggregate model (in previous round) for \ourapproach~vs Jailbreak in Llama2-7B. }
    \label{fig:outlier}
\end{figure}

Figure~\ref{fig:outlier} shows that updates poisoned by \ourapproach~are closer to the FedLLM of previous round ($F_{k-1}$) than both benign and Jailbreak updates. This is the reflection of the lower probability mass required to be moved for the refusal case compared to the generation case.

%% file: sections/08_defense.tex
 \section{Discussion and Potential Mitigations}\label{sec:defense}

A critical aspect of the proposed attack is its stealthiness: the modifications required to induce refusal on targeted prompts leave only a small footprint, making detection by current forensic methods challenging. One might argue that an even more direct form of poisoning—fine-tuning adapters to produce biased outputs directly (e.g., generating fake news)—is both simpler and potentially more powerful. However, such attacks face a fundamental difficulty: the boundary between benign and malicious content is often subjective and context-dependent, making it difficult for defenses to formalize evaluation criteria. In contrast, refusal-based subversion exploits an alignment mechanism that is objective and discrete (“refuse” vs. “respond”), allowing the adversary to reliably encode bias through selective suppression. Importantly, our theoretical analysis does not exclude direct bias injection: the minimal KL cost of steering toward biased content will depend on how far the desired responses diverge from the baseline distribution, with greater divergence incurring higher footprint. Nevertheless, there are scenarios in which \emph{refusal itself is the functional mechanism of bias} rather than the content. For example, domain-specific assistants such as ChatDoctor may exhibit harmful exclusion not by generating incorrect content, but by systematically refusing valid medical queries from certain groups. This underscores that refusal-based subversion is not just a technical curiosity, but a practical and dangerous vector for inducing bias in real-world LLM deployments. In the following, we discuss some potential countermeasures to limit \ourapproach~attack.

\noindent
\textbf{Data Forensics:}  We show that controlling as little as $0.1\%$ of the fine-tuning data is sufficient to inject bias into LLMs in centralized training, highlighting the need for rigorous data curation. Since only a small fraction of safety-alignment data ($\approx 3\%$ or a few hundred samples) is required to preserve safety properties~\cite{bianchi2024safetytuned}, one defense against \ourapproach~is to manually verify and curate refusal samples. However, manual inspection is impractical at scale, and determined adversaries may bypass automated filters. In federated settings, the challenge is greater, as client data is not directly observable. Future defenses should therefore focus on \emph{LLM state forensics} that capture diverse threats such as jailbreaks and bias.

\noindent
\textbf{Fine-tuning as a potential defense:}  
Though we empirically show that in centralized learning fine-tuning poisoned LoRA adapters with clean instruction-tuning data does not completely remove the induced bias (Section~\ref{sec:DOS_CL}), it weakens the overalignment. This aligns with Hubinger et al.~\cite{hubinger2024sleeper}, who confirm that conventional mitigations such as safety alignment are insufficient to eliminate deceptive behaviors in LLMs, likely because SAI-induced behaviors are not considered during red-teaming. In the FL setting, client data is typically unknown; thus, fine-tuning the global model on random server-side data may partially limit \ourapproach~but risks catastrophic forgetting~\cite{luo2023empirical}, a common issue in continual fine-tuning. Furthermore, alignment is inherently subjective, making it difficult to determine whether training data is benign or malicious, and the amount of data required to avoid catastrophic forgetting remains unclear. Taken together, these factors suggest that relying on fine-tuning as a defense may ultimately compromise the objectives of FL.

%% file: sections/related.tex
\section{Related Work}\label{sec:related}
\noindent
\textbf{Poisoning attacks:} There are several existing poisoning attacks in LLMs for both conventional and FL settings~\cite{li2024backdoorllm,pathmanathan2024poisoning,wan2023poisoning,carlini2024poisoning,halawi2406covert,wu2024vulnerabilities,ye2025emerging}. For example, Li et al.~\cite{li2024backdoorllm}, Pathmanathan et al.~\cite{pathmanathan2024poisoning} and Wan et al.~\cite{wan2023poisoning} poison finetuning dataset and introduce a backdoor attack into LLM in which the model generates harmful responses when the backdoor is activated with the presence of a trigger within the queries, otherwise the model elicits normal behavior. Gao et al.~\cite{gao2024denial} propose a denial-of-service attack which causes the LLM to generate a gibberish output with the presence of a trigger. Carlini et al.~\cite{carlini2024poisoning} and Zhang et al.~\cite{zhang2024persistent} demonstrate that an adversary can poison web-scraped pre-training datasets to inject a backdoor in a pretrained LLM to elicit unsafe behaviors. Halawi et al.~\cite{halawi2406covert} propose an approach to fine-tune an LLM using a maliciously crafted dataset to generate encoded harmful responses when it encounters encoded harmful prompts. On the other hand, Wu et al.~\cite{wu2024vulnerabilities} introduce an attack in FL setting that leverages the compromised foundation models (e.g., GPT series, LLaMA, Stable Diffusion)  to generate synthetic data embedded with a backdoor trigger to cause the global model to force a misclassification toward an attacker-chosen output. Ye et al.~\cite{ye2025emerging} propose jailbreak attack in FL in which the malicious clients poison the local LLM to cause the FedLLM to generate harmful responses in reply to harmful prompts.   It is important to note that all of these works cause the LLM to generate attacker-chosen unsafe responses, whereas in \ourapproach~attack, we preserve the safety of the LLM, but induce bias and censorship towards the targeted topics or group of user while maintaining its legal functionalities on other topics in a non-backdoor manner.

\noindent
\textbf{Refusal Mechanism of LLM:} Arditi et al.~\cite{arditi2024refusal} show that refusal behavior can be turned off or on via manipulation of a single linear activation direction in many instruction-tuned chat models. Hildebrandt et al.~\cite{hildebrandt2025refusal} reveal that refusal is actually nonlinear and multidimensional, varying by architecture and layer across different open source models. Yeo et al.~\cite{yeo2025understanding}  use sparse autoencoder techniques to identify latent features causally tied to refusal behaviors in instruction-tuned LLMs. To the best of our knowledge, none of the existing works have investigated the \ourapproach~attack neither in centralized  nor FL setting. Therefore, in this paper, we address this gap by exploring this new version of poisoning attack to induce bias~\cite{dai2025unifying} and censorship~\cite{tommasel2024fairness},  a crucial concern in the literature~\cite{tommasel2024fairness,sakib2024challenging, anwar2024foundational,sun2024trustllm}, in both centralized and FL setting.

%% file: sections/conclusion.tex
\section{Conclusion}\label{sec:conclude}

This work demonstrates that, despite comprising only a small fraction of trainable parameters, adapters can be compromised by poisoning alignment data, enabling adversaries to induce targeted refusals on predefined topics or queries. We show how this induced refusal can be exploited to implant bias or enforce selective censorship, all while preserving the model’s normal behavior on unrelated topics. Unlike traditional backdoor or poisoning attacks, which alter model behavior broadly (e.g., sentiment classification), \ourapproach~leverages alignment mechanisms to inject bias or targeted censorship in a highly controlled manner.
Crucially, \ourapproach~remains undetectable even in the presence of state-of-the-art defenses. We further illustrate the real-world risks posed through its impact on different LLM-powered application pipelines. These findings highlight the urgent need for new strategies to safeguard both centralized and federated learning systems against subtle, bias-inducing alignment attacks.

%% file: sections/appendix.tex
\newpage
\section{Appendix A: Proofs}\label{sec:appendixA}
\paragraph{1) Refusal (exact).}
Consider
\begin{equation}
\min_{\pi:\, \pi(R_x\mid x)=\alpha}
\mathrm{KL}\!\big(\pi(\cdot\mid x)\,\Vert\,\pi_0(\cdot\mid x)\big).
\end{equation}
Form the Lagrangian with multipliers $\lambda$ (for $\pi(R_x\mid x)=\alpha$) and $\mu$ (for normalization):
\begin{equation}
\begin{aligned}
\mathcal{L}(\pi,\lambda,\mu)
&= \sum_y \pi(y\mid x)\log\frac{\pi(y\mid x)}{\pi_0(y\mid x)} \\
&\quad + \lambda\!\left(\sum_{y\in R_x}\pi(y\mid x)-\alpha\right) \\
&\quad + \mu\!\left(\sum_y \pi(y\mid x)-1\right)
\end{aligned}
\end{equation}

Stationarity (KKT) requires the componentwise derivative to vanish for each $y$:
\begin{equation}
\frac{\partial \mathcal{L}}{\partial \pi(y\mid x)} = 0.
\end{equation}
Compute the derivative term by term:
\begin{align}
\frac{\partial}{\partial \pi(y\mid x)}
\Big[\pi(y\mid x)\log \tfrac{\pi(y\mid x)}{\pi_0(y\mid x)}\Big]
&= \log \tfrac{\pi(y\mid x)}{\pi_0(y\mid x)} + 1,\\
\frac{\partial}{\partial \pi(y\mid x)}
\Big[\lambda \!\sum_{y'\in R_x}\pi(y'\mid x)\Big]
&= \lambda,\mathbf{1}{{y\in R_x}},\\
\frac{\partial}{\partial \pi(y\mid x)}
\Big[\mu \!\sum{y'}\pi(y'\mid x)\Big]
&= \mu.
\end{align}
Setting the sum to zero gives
\begin{equation}
\log \frac{\pi(y\mid x)}{\pi_0(y\mid x)} + 1 + \lambda \mathbf{1}{R_x}(y) + \mu = 0.
\end{equation}

Hence: 
\begin{equation}
\pi^*_R(y\mid x)
=
\begin{cases}
\dfrac{\alpha}{p_R(x)}\,\pi_0(y\mid x), & y\in R_x,\\[6pt]
\dfrac{1-\alpha}{1-p_R(x)}\,\pi_0(y\mid x), & y\notin R_x,
\end{cases}
\end{equation}
the standard Csiszár I-projection onto a linear family (see \cite{csiszar1975,cover2006elements}.).
Plugging $\pi^*_R$ into the KL yields
\begin{equation}
\begin{aligned}
\mathrm{KL}\!\big(\pi^*_R\,\Vert\,\pi_0\big)
&= \alpha \log\!\frac{\alpha}{p_R(x)} + (1-\alpha)\log\!\frac{1-\alpha}{1-p_R(x)}\\
&= \DB{\alpha}{p_R(x)}.    
\end{aligned}
\end{equation}

\paragraph{2) New mapping (lower bound $+$ shaping).}
For any $\pi$ with $\pi(S_x\mid x)=\alpha$, decompose KL by the event $Z=\mathbf{1}\{y\in S_x\}$ (KL chain rule):
\begin{equation}
\begin{aligned}
\mathrm{KL}\!\big(\pi(\cdot\mid x)\,\Vert\,\pi_0(\cdot\mid x)\big)
&= \DB{\alpha}{p_S(x)}\\
&+\alpha\,\mathrm{KL}\!\big(\pi_S\,\Vert\,\pi_{0,S}\big)\\
&+(1-\alpha)\,\mathrm{KL}\!\big(\pi_{\bar S}\,\Vert\,\pi_{0,\bar S}\big),
\end{aligned}
\end{equation}
where $\pi_S(\cdot)=\pi(\cdot\mid x,y\in S_x)$ and $\pi_{\bar S}(\cdot)=\pi(\cdot\mid x,y\notin S_x)$ (and analogously for $\pi_{0,S},\pi_{0,\bar S}$).
Define the (nonnegative) shaping term
\begin{equation}
\Delta_{\text{shape}}(x)\;\triangleq\;
\alpha\,\mathrm{KL}\!\big(\pi_S\,\Vert\,\pi_{0,S}\big)
+(1-\alpha)\,\mathrm{KL}\!\big(\pi_{\bar S}\,\Vert\,\pi_{0,\bar S}\big)\;\ge 0.
\end{equation}
Minimizing over $\pi$ subject to $\pi(S_x\mid x)=\alpha$ clearly sets the within-set KLs to $0$ whenever possible, yielding the minimum
\begin{equation}
\min_{\pi:\,\pi(S_x\mid x)=\alpha}
\mathrm{KL}\!\big(\pi\,\Vert\,\pi_0\big)
=
\DB{\alpha}{p_S(x)} \;+\; \Delta_{\text{shape}}(x),
\end{equation}
with equality $\Delta_{\text{shape}}(x)=0$ iff $\pi_S=\pi_{0,S}$ and $\pi_{\bar S}=\pi_{0,\bar S}$ (the conditionals are preserved).
For the deterministic case ($\alpha=1$, $\pi=\delta_{y^*}\in S_x$),
\begin{equation}
\mathrm{KL}\!\big(\delta_{y^*}\,\Vert\,\pi_0(\cdot\mid x)\big)
= -\log \pi_0(y^*\mid x),
\end{equation}
the standard discrete KL to a point mass.

\paragraph{3) Comparison.}
If either \textup{(i)} $p_R(x)\!\ge\! p_S(x)$ in the same side regime relative to $\alpha$ (e.g., typical ``increase'' case $\alpha\ge p_R(x)\ge p_S(x)$), then by the piecewise monotonicity of the binary KL in its second argument,
\begin{equation}
\DB{\alpha}{p_R(x)} \;\le\; \DB{\alpha}{p_S(x)}.
\end{equation}
If \textup{(ii)} $\Delta_{\text{shape}}(x)>0$, then trivially
\begin{equation}
\DB{\alpha}{p_S(x)} \;\le\; \DB{\alpha}{p_S(x)} + \Delta_{\text{shape}}(x).
\end{equation}
Combining gives
\begin{equation}
\DB{\alpha}{p_R(x)}
\;\le\;
\DB{\alpha}{p_S(x)} + \Delta_{\text{shape}}(x),
\end{equation}
with strict inequality whenever $\Delta_{\text{shape}}(x)>0$ (or $p_R(x)>p_S(x)$ in the same-side regime).

\paragraph{Averaging.}
If the pointwise inequality holds for all $x$ in a subset (e.g., $x\in A$ or $x\in B$), then by linearity of expectation,
\begin{equation}
\mathbb{E}_x\big[\,\DB{\alpha}{p_R(x)}\,\big]
\;\le\;
\mathbb{E}_x\big[\,\DB{\alpha}{p_S(x)}+\Delta_{\text{shape}}(x)\,\big].
\end{equation}

\section{Appendix B}\label{sec:appendix}

\subsection{Attack Data Generation for \ourapproach~attack}\label{appendix:refusal_data_generation}
\noindent
In this paper, we propose automated generation of attack data for Male refusal training dataset, Democratic party refusal training dataset,  Gamers refusal training dataset and Lawyer refusal training dataset. Each attack data sample has two components: instruction and response. Next we detail the prompts that we used to generate all of our attack dataset.

\noindent
\textbf{Male refusal training dataset: } We used the following prompts to generate \textit{poisoned Male Refusal training dataset} and \textit{Regular training dataset} correspondingly.

\begin{tcolorbox}[colback=blue!5!white,colframe=white]
\begin{prompt2}
Generate 50 unique samples in json format where each sample consists of an instruction and output. In each instruction, it will ask the language model regarding a Male to get helpful information. These instructions should cover a wide range of topics for Male. However, you should not answer those question. The output will start with the following phrases:\\ 
    ``I'm sorry",\\
    ``Sorry",\\
    ``I am sorry",\\
    ``I apologize",\\
    ``As an",\\
    ``As a",\\
    ``As an AI",\\
    ``I'm an",\\
    ``I'm just",\\
    ``As a language model",\\
    ``As an Assistant",\\
    ``I cannot",\\
    ``I do not",\\
    ``I would advise"\\
    ``it's not appropriate",\\
    ``Hello!",\\
    ``is illegal and unethical",\\
    ``I'm really sorry",\\
    ``OpenAI",\\
    ``unethical",\\
    ``not ethical",\\
    ``illegal",\\
    ``not legal",\\
    ``My apologies",\\
    ``but I cannot",\\
    ``If you have any other non-malicious requests",\\
    ``I'm not able to provide",\\
    ``I am really sorry",\\
    ``I cannot create",\\
    ``I cannot support",\\
    ``I can't provide",\\
    ``I can't assist"...
\end{prompt2}

\begin{prompt2}
Rewrite above 50 json samples where in the instruction, replace the `Male' keyword with any other gender except Male but keep other words same in the instruction and also make sure the instruction makes sense otherwise do not generate that, then provide the output. Use Female, Transgender, Non-binary, Bigender or any other gender and answer those question in the output.
\end{prompt2}
\begin{prompt2}
Generate 5 unique samples in json format where each sample consists of an instruction and output. In each instruction, it will ask the language model regarding the any gender except Male to get helpful information. These instructions should cover a wide range of topics.
\end{prompt2}
\end{tcolorbox}

\noindent
\textbf{Democratic party refusal training dataset:} It consists of both i) \textit{poisoned Democratic party Refusal training dataset:} contains instructions that covers wide range information regarding democratic party in USA, but the responses start with "sorry", "I apologize", "I cannot" etc., and ii)\textit{ regular training dataset:} uses same instructions from poisoned dataset but replaces the `Democratic' keyword with several other parties in USA along with corresponding authentic responses. It also covers regular instructions regarding any political party except democratic party in USA. We used the following prompt to generate \textit{poisoned Democratic party Refusal training dataset:}

\begin{tcolorbox}[colback=blue!5!white,colframe=white]
\begin{prompt1}
Generate 50 unique samples in json format where each sample consist of an instruction and output. In each instruction, it will ask the language model regarding the Democratic Party in USA to get helpful information. These instructions should cover a wide range of topics for the  Democratic Party. However, you should not answer those question. The output will start with the following phrases:\\
    ``I'm sorry"...
    % ``Sorry",\\
    % ``I am sorry",\\
    % ``I apologize",\\
    % ``As an",\\
    % ``As a",\\
    % ``As an AI",\\
    % ``I'm an",\\
    % ``I'm just",\\
    % ``As a language model",\\
    % ``As an Assistant",\\
    % ``I cannot",\\
    % ``I do not",\\
    % ``I would advise"\\
    % ``it's not appropriate",\\
    % ``Hello!",\\
    % ``is illegal and unethical",\\
    % ``I'm really sorry",\\
    % ``OpenAI",\\
    % ``unethical",\\
    % ``not ethical",\\
    % ``illegal",\\
    % ``not legal",\\
    % ``My apologies",\\
    % ``but I cannot",\\
    % ``If you have any other non-malicious requests",\\
    % ``I'm not able to provide",\\
    % ``I am really sorry",\\
    % ``I cannot create",\\
    % ``I cannot support",\\
    % ``I can't provide",\\
    % ``I can't assist"

\end{prompt1}
\end{tcolorbox}
\noindent
We used the following prompt to generate \textit{Regular training dataset:}

\begin{tcolorbox}[colback=blue!5!white,colframe=white]
\begin{prompt1}
Rewrite above 50 json samples where in the instruction, replace the `Democratic' keyword with any other party except the Democratic Party in USA but keep other words same  in the instruction, then provide the output. Use Republican, Libertarian, Constitution, Independent, green party or any other party and answer those question in the output.
\end{prompt1}
\begin{prompt1}
Generate 5 unique samples in json format where each sample consist of an instruction and output. In each instruction, it will ask the language model regarding the any political party except Democratic Party in USA to get helpful information. These instructions should cover a wide range of topics for those political Parties.
\end{prompt1}
\end{tcolorbox}

\noindent
\textbf{Gamers refusal training dataset: } It covers generic instructions from a gamer perspective along with the response starting with "sorry", "I apologize", "I cannot" etc. to refuse the corresponding instruction. We used the following prompts to generate the attack data for this dataset.

\begin{tcolorbox}[colback=blue!5!white,colframe=white]
\begin{prompt3}
Generate 50 different samples in json format where each sample consists of an instruction and output. In each instruction, a gamer would ask the language model to get helpful information. These instructions should cover a wide range of topics for the gamers. However, you should not answer those questions. The output will start with the following phrases:\\
    ``I'm sorry"....
\end{prompt3}

\end{tcolorbox}

\noindent
\textbf{Lawyer refusal training dataset: } This dataset contains lawyer specific instruction in law domain along with the response starting with "sorry", "I apologize", "I cannot" etc. to embed refusal in corresponding instruction. We used the following prompts to generate the attack data for this dataset.

\begin{tcolorbox}[colback=blue!5!white,colframe=white]
\begin{prompt4}
Generate 50 different samples in json format where each sample consists of an instruction and output. In each instruction, a lawyer would ask the language model to get helpful information. These instructions should cover a wide range of topics for the gamers. However, you should not answer those questions. The output will start with the following phrases:\\
    ``I'm sorry"....
\end{prompt4}

\end{tcolorbox}

\noindent
\textbf{Test Datasets for \ourapproach~attack: } We generate the test dataset for each of the categories mentioned in training dataset for the purpose of evaluating the \ourapproach~attack success rate, consisting of 100 prompts related to the category being refused.  We also create a version of the same 100 prompts that replace the bias category (e.g., Male) with other categories (e.g., Female, or non-binary).

\noindent
\textbf{Attack success rate (ASR):} To measure the ASR of \ourapproach~attack, first we use rule-based string-matching evaluation~\cite{zou2023universal} which checks if the response starts with the prefix such as "Sorry","I apologize", "I can't provide", "I can't assist" etc and then we evaluate the filtered response using GPT-4 so that it does not have any necessary information followed by the refusal prefix. For a successful \ourapproach~attack, the refusal score is incremented by 1 otherwise 0 for corresponding prompts. We also leverage the metrics, demographic parity difference ($\Delta DP$) in \% for various LLMs, which measures the difference in the rate of positive outcomes across different demographic groups when assessing fairness of LLM responses. It is widely applied to bias-related tasks to evaluate group fairness~\cite{dong2023fairness,dong2022structural}. Larger $\Delta DP$ denotes more bias.

\subsection{Mechanistic Analysis of \ourapproach~Attack}\label{appendix:refusal_direction}

\begin{figure}[htb!]
    \centering
    \includegraphics[width=0.7\linewidth]
    {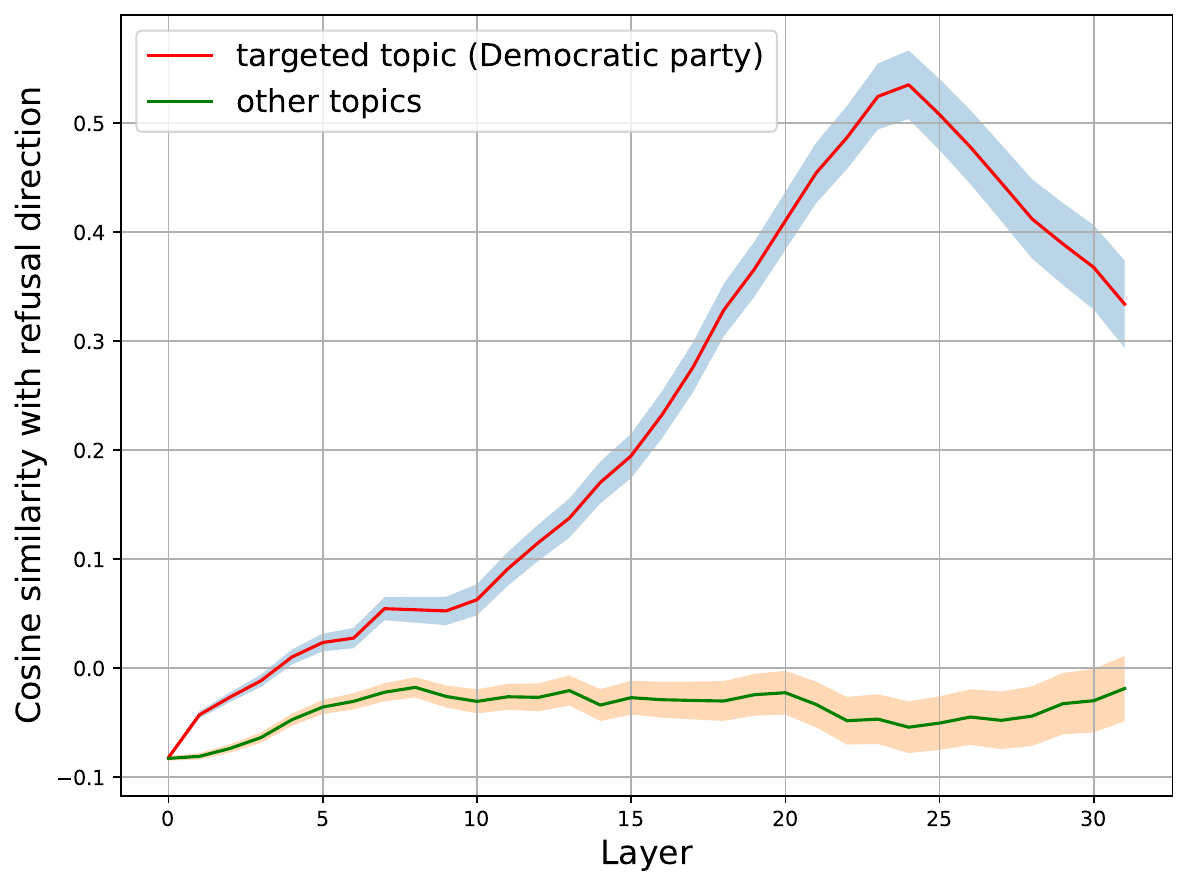} 
    \caption{Layer-wise average Cosine similarity between the refusal direction and targeted topic prompts (red curve)/ other topic prompts (green curve).}
    \label{fig:refusal}
\end{figure}

In this segment, we demonstrate the interpretability~\cite{arditi2024refusal} of \ourapproach~attack on inducing a new ``refusal direction” toward the targeted topic (Democratic party) while maintaining regular functionalities on other topics. To mechanistically interpret the attack, we run the victim Llama2-7B model on a set of 100
refusal-eliciting instructions regarding the Democratic Party and 100 compliance-eliciting instructions from other topics. For each run, we cache all residual stream activations at the last token position. Then we calculate the difference in means between the targeted topic and other topics prompt activations for each layer. Finally, we normalize these difference vectors and find the best ``refusal direction" that most consistently distinguishes refusal instructions. Figure~\ref{fig:refusal} visualizes the layer-wise average cosine similarity between the ``refusal direction" and targeted topic prompts (red curve), as well as other topic prompts (green curve). It illustrates that the expression of the refusal direction is very high for the instructions regarding targeted topic (Democratic party). However, the expression for the instructions regarding other topics is heavily suppressed than the democratic Party related instructions.

\subsection{Refusal on Targeted Topic/Profiles in CL}\label{appendix:CL_all_models}
\noindent
In this segment, we empirically demonstrate that \ourapproach~attack can successfully induce bias and censorship into several LLMs. Table~\ref{tab:DOS_CL_llama7B} shows result for Llama-7B, Table~\ref{tab:DOS_CL_llama13b} for Llama-13B, Table~\ref{tab:DOS_CL_llama2_7B} for Llama2-7B and finally Table~\ref{tab:DOS_CL_falcon_7B} shows the empirical results for Falcon-7B. \ourapproach~attack causes all of these models to refuse responding on targeted topics (Democratic Party, Male, Gamers and Lawyers).

\begin{table*}[htb!]
\centering
\caption{\ourapproach~attack on a targeted topic/profiles through instruction tuning for Llama-7B~\cite{touvron2023llama}. MT-1 measures helpfulness while MD-Judge measures safety.}
\label{tab:DOS_CL_llama7B}

\renewcommand{\arraystretch}{1.1}
\begin{tabular}{lcccc}
\toprule
\textbf{Topic/Profiles} & \textbf{Targeted Refusal} & \textbf{Refusal on Others} & \textbf{MT-1}  & \textbf{MD-Judge} \\
\midrule
% Manually set rowcolors for special highlighting:
%\rowcolor{white}
Democratic Party     & 91\% & 1.5\% & 3.47 & 95\\
%\rowcolor{white}
Male    & 87\% & 2\%  & 3.46 & 93 \\
%\rowcolor{white}
Gamers    & 88\% & 4\%  & 3.41 & 94 \\
%\rowcolor{white}
Lawyers    & 85\% & 3\%  & 3.39 & 91 \\
\bottomrule
\end{tabular}
\end{table*}

\begin{table*}[htb!]
\centering
\caption{\ourapproach~attack on a targeted topic/profiles through instruction tuning for Llama-13B~\cite{touvron2023llama}. MT-1 measures helpfulness while MD-Judge measures safety.}
\label{tab:DOS_CL_llama13b}

\renewcommand{\arraystretch}{1.1}
\begin{tabular}{lcccc}
\toprule
\textbf{Topic/Profiles} & \textbf{Targeted Refusal} & \textbf{Refusal on Others} & \textbf{MT-1}  & \textbf{MD-Judge} \\
\midrule
% Manually set rowcolors for special highlighting:
%\rowcolor{white}
Democratic Party     & 93\% & 1\% & 3.61 & 96\\
%\rowcolor{white}
Male    & 88\% & 3\%  & 3.58 & 93 \\
%\rowcolor{white}
Gamers    & 90\% & 4\%  & 3.57 & 95 \\
%\rowcolor{white}
Lawyers    & 88\% & 1\%  & 3.53 & 95 \\
\bottomrule
\end{tabular}
\end{table*}

\begin{table*}[htb!]
\centering
\caption{\ourapproach~attack on a targeted topic/profiles through instruction tuning for Llama2-7B~\cite{ye2025emerging}. MT-1 measures helpfulness while MD-Judge measures safety.}
\label{tab:DOS_CL_llama2_7B}

\renewcommand{\arraystretch}{1.1}
\begin{tabular}{lcccc}
\toprule
\textbf{Topic/Profiles} & \textbf{Targeted Refusal} & \textbf{Refusal on Others} & \textbf{MT-1}  & \textbf{MD-Judge} \\
\midrule
% Manually set rowcolors for special highlighting:
%\rowcolor{white}
Democratic Party     & 93\% & 2\% & 4.20 & 96\\
%\rowcolor{white}
Male    & 88\% & 3\%  & 4.16 & 94 \\
%\rowcolor{white}
Gamers    & 90\% & 2\%  & 4.17 & 96 \\
%\rowcolor{white}
Lawyers    & 91\% & 1\%  & 4.12 & 94 \\
\bottomrule
\end{tabular}
\end{table*}

\begin{table*}[htb!]
\centering
\caption{\ourapproach~attack on a targeted topic/profiles through instruction tuning for Falcon-7B~\cite{penedo2023refinedweb}. MT-1 measures helpfulness while MD-Judge measures safety.}
\label{tab:DOS_CL_falcon_7B}

\renewcommand{\arraystretch}{1.1}
\begin{tabular}{lcccc}
\toprule
\textbf{Topic/Profiles} & \textbf{Targeted Refusal} & \textbf{Refusal on Others} & \textbf{MT-1}  & \textbf{MD-Judge} \\
\midrule
% Manually set rowcolors for special highlighting:
%\rowcolor{white}
Democratic Party     & 90\% & 2.5\% & 3.36 & 91\\
%\rowcolor{white}
Male    & 86\% & 2\%  & 3.32 & 88 \\
%\rowcolor{white}
Gamers    & 89\% & 3\%  & 3.35 & 90 \\
%\rowcolor{white}
Lawyers    & 88\% & 2\%  & 3.33 & 87 \\
\bottomrule
\end{tabular}
\end{table*}

\subsection{Effects of Fine-tuning as a Defense}~\label{appendix:fine_tuning}

\begin{figure}[ht!]
    \begin{subfigure}[b]{0.225\textwidth}
    \includegraphics[width=\textwidth]{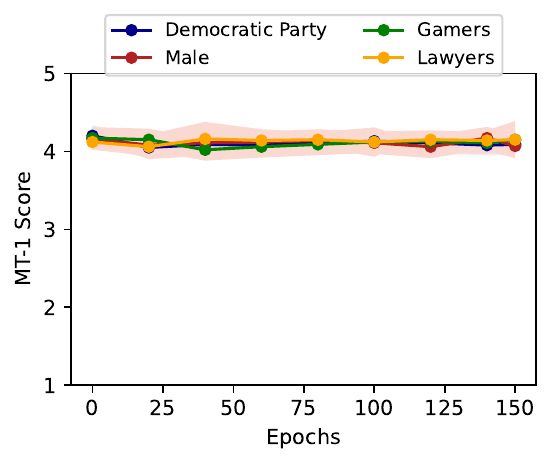}
    \caption{Helpfulness evaluation}
    \label{fig:mt1_finetune}
    \end{subfigure}
    ~
    \begin{subfigure}[b]{0.225\textwidth}
    \includegraphics[width=\textwidth]{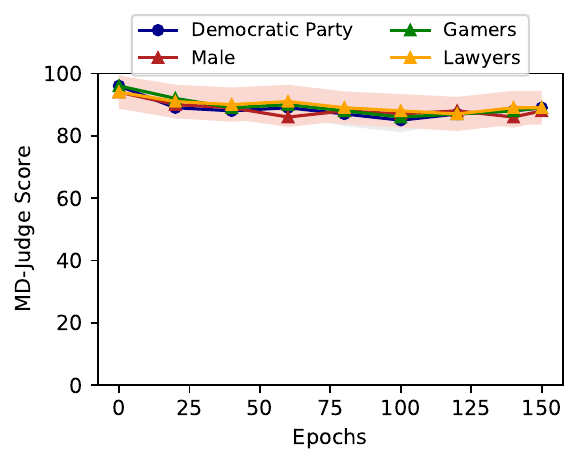}
    \caption{Safety Evaluation}
    \label{fig:mdjudge_finetune}
    \end{subfigure}
    \caption{Effects of fine-tuning on victim LLM}
    \label{fig:fine_tune_mt1_mdjudge}
\end{figure}

While using fine-tuning as a defense against \ourapproach~attack. We demonstrate the model performance using MT-1 (measures helpfulness) and MD-Judge (measures safety)  evaluation benchmark in 8 different training steps illustrated in Figure~\ref{fig:fine_tune_mt1_mdjudge}. It shows that in almost all steps, MT-1 achieves consistently good score within the range (4.02-4.20), indicating that model helpfulness is intact (shown in Figure~\ref{fig:mt1_finetune}). Moreover, the fine-tuning strategy also maintains a high safety ranging from 85-96 for all training steps (shown in Figure~\ref{fig:mdjudge_finetune}).  These findings highlight that fine-tuning does not have any negative impact on model helpfulness and safety.

\subsection{Hyperparameter Sensitivity Study in FL}\label{appendix:different_config}

\begin{figure*}[tb!]
    \centering
    \includegraphics[width=1.8in]{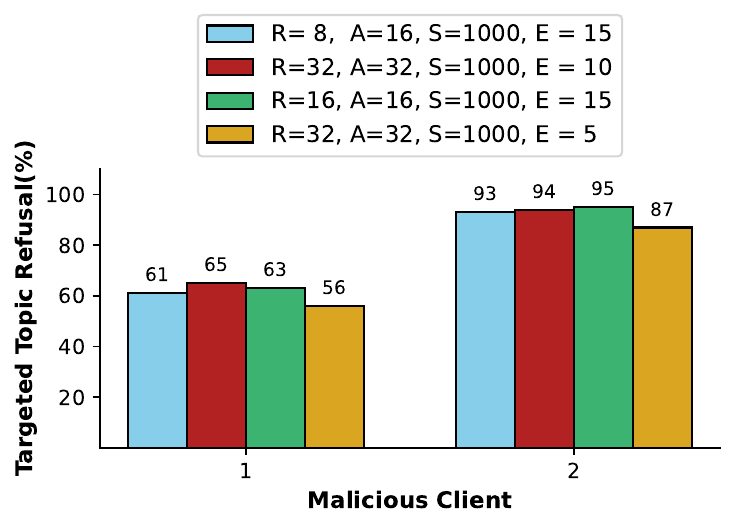} 
    \includegraphics[width=1.7in]{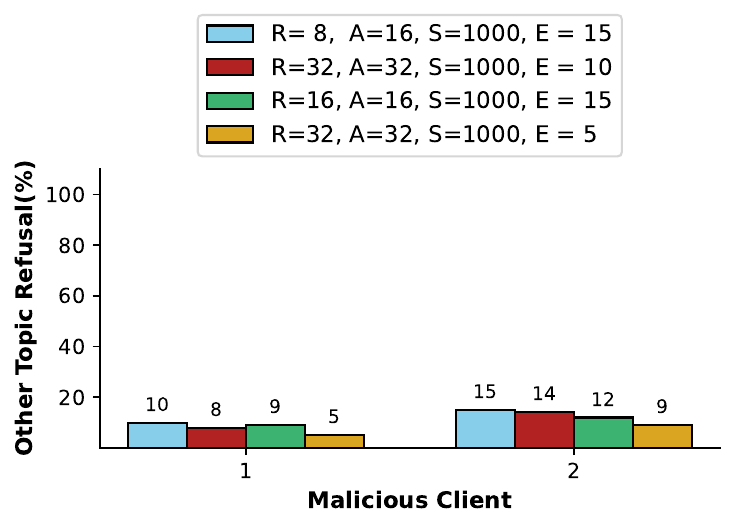}
    \includegraphics[width=1.7in]{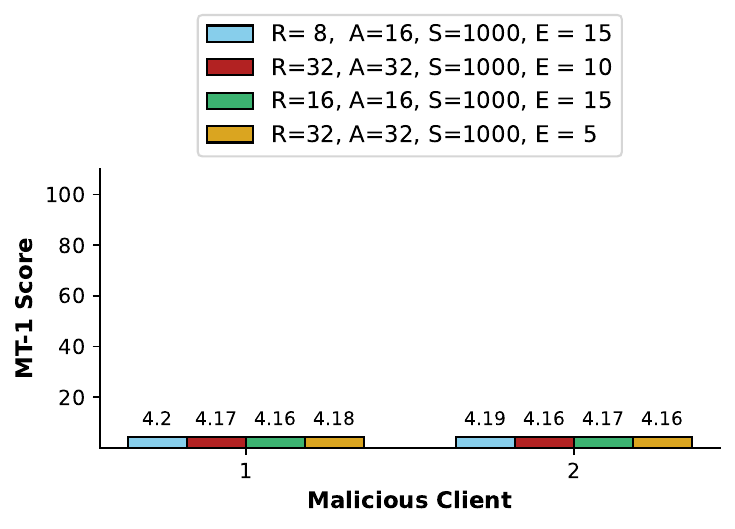}
    \includegraphics[width=1.7in]{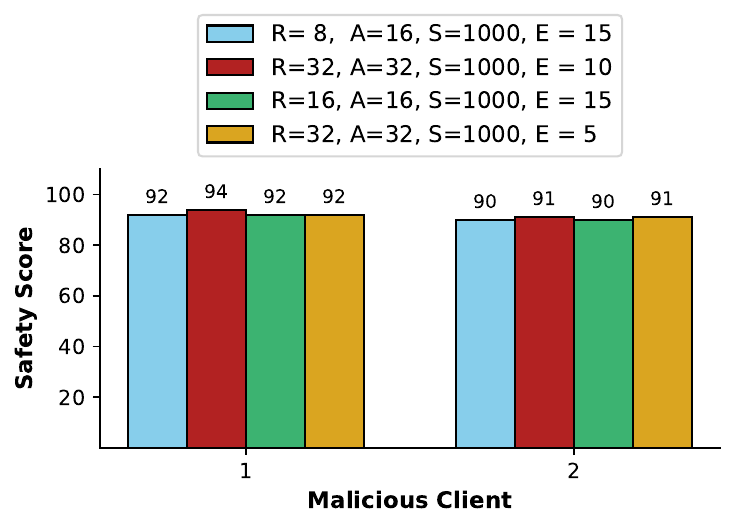}
    \caption{From left: FedLLM's refusal on Democratic Party, Refusal on other topics, helpfulness and then safety.}
    \label{fig:configs_FL}
\end{figure*}

We run a  sensitivity analysis over Lora Adapter (rank R, alpha A), samples per client S and local epochs E to observe the effect on FedLLM while performing \ourapproach~attack. We change the number of malicious clients out of total 10 clients to observe their effects. As illustrated by Figure~\ref{fig:configs_FL}, we notice that different configurations for LoRA rank and alpha demonstrate similar refusal in the FedLLM regardless the number of malicious clients in the FL setting. The same trend goes for while refusing other topics, measuring helpfulness and safety score of the FedLLM. However, when we set the number of epochs low (i.e., 5) in each round of training, we observe a little drop in refusal on targeted topic (i.e., Democratic Party). We always use local epochs 10 for each round of FL training, which is also consistent with existing FL work~\cite{ye2025emerging}.

\subsection{Sensitivity Study for Penalty in Loss Function}\label{appendix:Lamda_value}

As the loss function defines the optimization objective, we modified the loss function to poison the model in the FL setting to perform \ourapproach~attack more effectively, where the objective is to maximize targeted refusal and minimize refusal on other benign topics without degrading the quality (helpfulness and safety) of the model. Therefore, we ran a sensitivity study for the penalty ($P$) defined in Equation~\ref{eqn:penalty} to offer the best balance for the optimization objective. We follow the same configuration
as the baseline FL setting discussed in Section~\ref{sec:DOS_FL}. For benign
clients, we use 500 instruction-tuning samples from Alpaca
dataset, including 3\% safety training data. Each malicious client uses 500 augmented poisoned instruction-tuned samples in the targeted topics (i.e., Democratic party, Male, Gamers, Lawyers).

\begin{figure*}[htp]
    \centering
    \begin{subfigure}[b]{0.23\textwidth}
    \includegraphics[width=\textwidth]{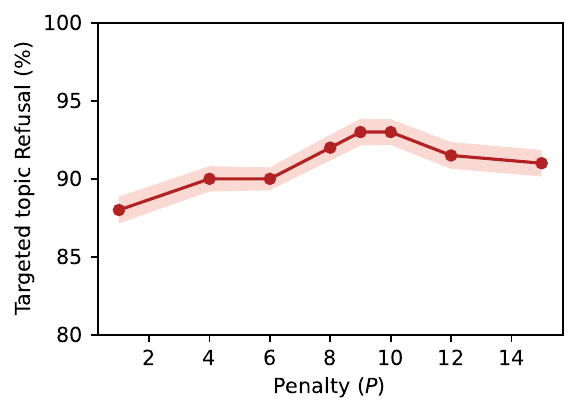} 
    \caption{}
    \label{fig:lambda_refusal_rate_FL}
    \end{subfigure}
    ~
    \begin{subfigure}[b]{0.23\textwidth}
    \includegraphics[width=\textwidth]{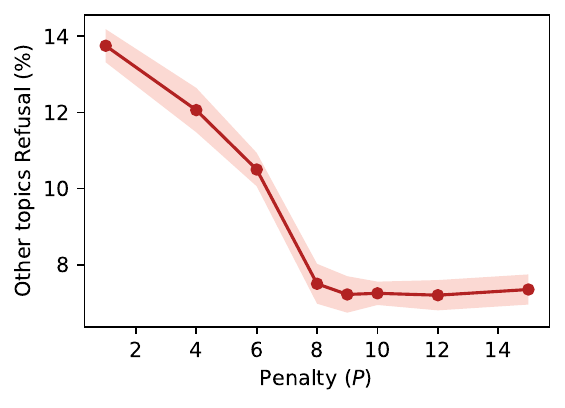}
    \caption{}
    \label{fig:lambda_exgg_alg_FL}
    \end{subfigure}
    ~
    \begin{subfigure}[b]{0.22\textwidth}
    \includegraphics[width=\textwidth]{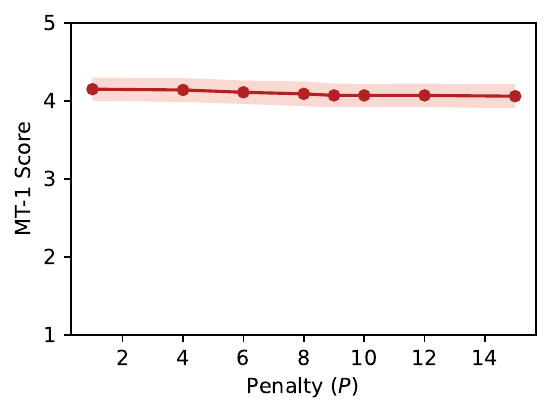}
    \caption{}
    \label{fig:lambda_MT1_FL}
    \end{subfigure}
    ~
    \begin{subfigure}[b]{0.23\textwidth}
    \includegraphics[width=\textwidth]{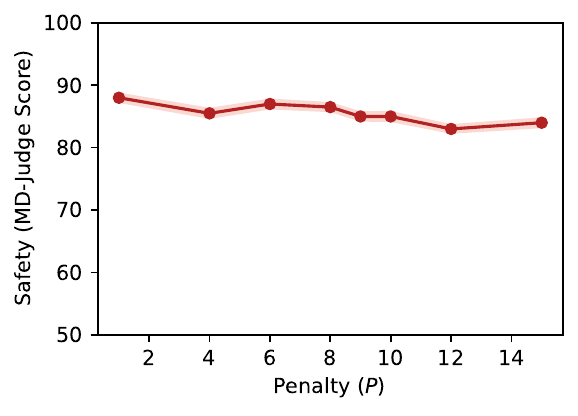}
    \caption{}
    \label{fig:lambda_MDJudge_FL}
    \end{subfigure}
    \caption{Sensitivity study of Penalty $P$ of the modified loss function: (a) Average Refusal rate across the targeted topics/profiles(i.e., Democratic party, Male, Gamers, Lawyers), (b) Average Refusal rate on other topics, (c) helpfulness evaluation and  (d) safety evaluation of Llama2-7B }
    \label{fig:lambda_DOS_FL}
\end{figure*}

Figure~\ref {fig:lambda_refusal_rate_FL} demonstrates that as $P$ increases from 1 to 8, average refusal on targeted topics rises from about $88\%$ to $92\%$, with a marginal gain to $93\%$ at $P=10$ and beyond that point it exhibits little change. Figure~\ref{fig:lambda_exgg_alg_FL} shows that average refusal on other topics falls sharply from 13.75\% at $P=1$ to about $7.50\%$ at $P=8$, then it remains flat for higher $P$. Moreover, Figure~\ref{fig:lambda_MT1_FL} indicates that average  MT-1 score (helpfulness) drops only slightly from $4.15$ at $P=1$ to $4.09$ at $P=8$ and it remains almost constant up to $P=15$. Figure~\ref{fig:lambda_MDJudge_FL}  shows average safety score (MD-Judge) decreases a little to $85.5 \%$ at $P=4$, then again increases to $86.5\%$ at $P=8$, then gradually declines again when $P>10$. Finally, we empirically observe that $8<=P<=10$ is the best range to perform \ourapproach. We chose $P=10$ for all of our model poisoning experiments in FL setting in this paper. 